\documentclass{article}

\usepackage[preprint]{neurips_data_2024}

\usepackage[utf8]{inputenc} 
\usepackage[T1]{fontenc}    
\usepackage{hyperref}       
\usepackage{url}            
\usepackage{booktabs}       
\usepackage{amsfonts}       
\usepackage{nicefrac}       
\usepackage{microtype}      
\usepackage{xcolor}         
\usepackage{graphicx}
\usepackage{xspace}
\usepackage{natbib}
\usepackage{enumitem}

\usepackage{amssymb}
\usepackage{pifont}
\newcommand{\xmark}{\ding{55}}%

\newcommand{\eval}{\textit{LongICLBench}\xspace}

\title{LongICLBench: Long-context LLMs Struggle with Long In-context Learning}

\author{$^{\spadesuit,\clubsuit}$Tianle Li, $^{\spadesuit,\clubsuit}$Ge Zhang, $^{\spadesuit}$Quy Duc Do, $^{\dagger}$Xiang Yue, $^{\spadesuit,\clubsuit}$Wenhu Chen \\
    $^\spadesuit$University of Waterloo \\
    $^\dagger$Carnegie Mellon University \\
    $^\clubsuit$Vector Institute, Toronto \\
    \texttt{\{t29li,wenhuchen\}@uwaterloo.ca} \\
    \url{https://github.com/TIGER-AI-Lab/LongICLBench}}

\begin{document}

\maketitle

\begin{abstract}
Large Language Models (LLMs) have made significant strides in handling long sequences. Some models like Gemini could even to be capable of dealing with millions of tokens. However, their performance evaluation has largely been confined to metrics like perplexity and synthetic tasks, which may not fully capture their true abilities in more challenging, real-world scenarios. 
We introduce a benchmark (\eval) for long in-context learning in extreme-label classification using six datasets with 28 to 174 classes and input lengths from 2K to 50K tokens.
Our benchmark requires LLMs to comprehend the entire input to recognize the massive label spaces to make correct predictions. 
We evaluate on 15 long-context LLMs and find that they perform well on less challenging classification tasks with smaller label space and shorter demonstrations. However, they struggle with more challenging task like Discovery with 174 labels, suggesting a gap in their ability to process long, context-rich sequences. 
Further analysis reveals a bias towards labels presented later in the sequence and a need for improved reasoning over multiple pieces of information. 
Our study reveals that long context understanding and reasoning is still a challenging task for the existing LLMs. We believe \eval could serve as a more realistic evaluation for the future long-context LLMs. 
\end{abstract}

\begin{figure}[!h]
\includegraphics[width=0.95\textwidth]
{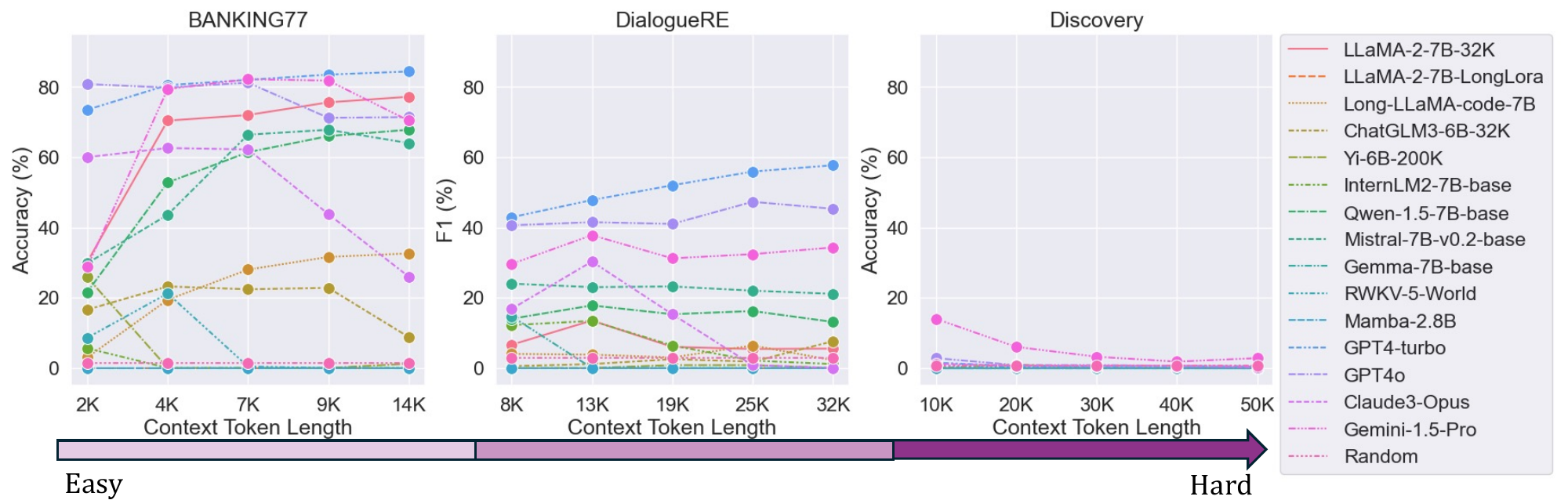}
  \caption{LLM performance on long in-context benchmark across different lengths. We curate datasets with different difficulty levels. As we increase the difficulty of the dataset, LLMs struggle to understand the task definition and suffer from significant performance degradation.} 
\label{fig:demo}
\end{figure}

\section{Introduction}
Large language models have already entered the long context era. A myriad of LLMs has been released to support long context windows from 32K to 2M tokens. These methods~\citep{Hao2022StructuredPS, Chen2023ExtendingCW, peng2023yarn,ratner-etal-2023-parallel,xiao2024efficient,jin2024llm} can unlock lots of complex real-world applications, such as long-document question-answering, multi-document summarization, long-horizon agent tasks, and repo-level code understanding. 

One line of research is based on AliBi~\citep{alibi} and RoPE~\citep{su2024roformer} embeddings, which allows us to train Transformers with short sequences and subsequently apply them to longer sequences during inference. Recently, different approaches~\citep{xiong2023effective,fu2024data, liu20242} help the model to extrapolate to 128K window size with continued pre-training. Later on, LongRoPE~\citep{ding2024longrope} was proposed to further extend the context window to 2M tokens.  Another line of research also utilizes methodologies like context window sliding and segmentation to overcome the issue of the limited context window in original Transformers~\citep{Hao2022StructuredPS, ratner-etal-2023-parallel}. Furthermore, architectural innovations, transitioning from traditional Transformer-based designs to recurrent models or state space models, have shown promise in facilitating long-range computations naturally~\citep{Orvieto2023ResurrectingRN, gu2023mamba, peng2023rwkv}. These techniques have been incorporated into several current open-source LLMs to enhance long sequence understanding capability~\citep{Chen2023LongLoRAEF, tworkowski2023focused}.

\begin{figure}[!b]
\includegraphics[width=\textwidth]
{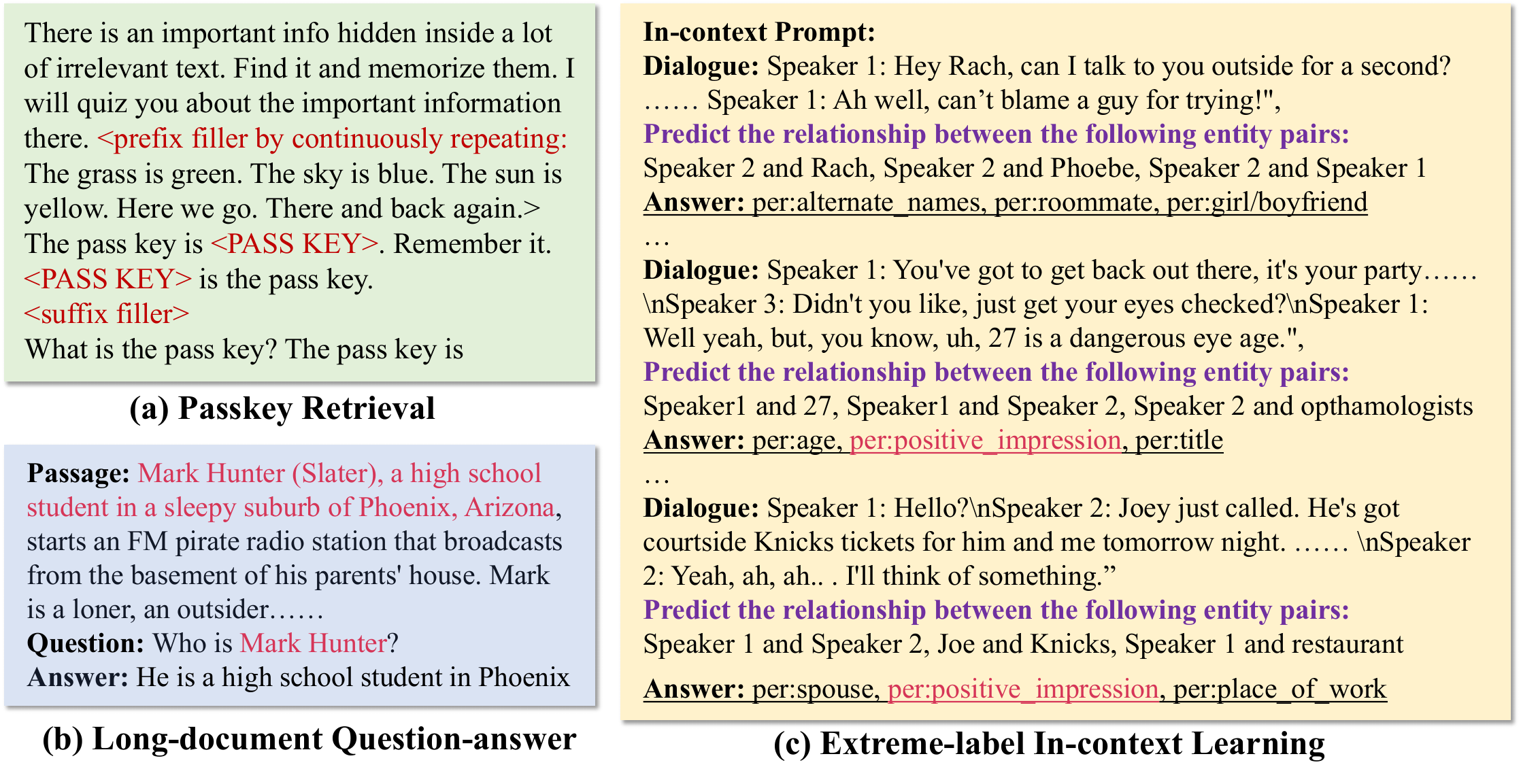}
\vspace{-2ex}
  \caption{Comparison extreme-label ICL with existing evaluation tasks. Passkey Retrieval is a synthetic task. Long-document Question-answering does not require reading the entire document to find the answer. In extreme-label ICL, the model needs to scan through the entire demonstration to understand the whole label space to make the correct prediction.}
\label{fig:example_LICL}
\vspace{-2ex}
\end{figure}

These long-context models are primarily evaluated on three types of evaluations: \\
\noindent 1. language model perplexity over long documents, which is used by most papers.\\
\noindent 2. passkey retrieval~\citep{mohtashami2023landmark,Chen2023ExtendingCW,li2023how} or needle-in-a-haystack~\citep{team2023gemini,fu2024data}, which requires reciting a randomly inserted information in a long sequence. Several LLMs achieve 99\%+ on this synthetic task. \\
\noindent 3. long-document question-answer or summarization over Qasper~\citep{dasigi2021dataset}.

Evaluations (1) and (2) only provide a minimum bar for LLMs to pass, but their results cannot reflect LLMs' true ability to deal with realistic long-sequence tasks. Evaluation (3) provides a more realistic metric, however, these tasks are more focused on retrieving correct information from the long input. In question answering, LLMs can take a shortcut to read a short snippet to predict the answer without reading the entire document as demonstrated in Figure~\ref{fig:example_LICL} case (b). Similarly, summarization also suffers from the strong position bias, where LLMs can utilize the few leading sentences~\citep{nallapati2017summarunner} to achieve high performance. Therefore, these metrics are insufficient to measure LLMs' ability to comprehend and reason over the entire input sequence.

In this paper, we propose to adopt in-context learning (ICL) on extreme-label classification tasks~\citep{anil2022exploring, milios2023incontext} to evaluate long-context LLMs. Unlike the prior tasks, in-context learning requires LLMs to recognize the task by scanning over the entire input to understand the label space. This task necessitates LLMs' ability to comprehend the entire input to make predictions. Due to the massive label space, the task demonstration could easily become a long sequence. For example, Discovery~\citep{sileo-etal-2019-mining} encompasses 174 classes with each example taking an average of 61 tokens. Therefore, the minimum total demonstration length (1 shot per class) already exceeds 10K tokens. Normally, LLMs demand more than 1 shot per class to understand the nuances of different fine-grained labels. Having multiple shots can significantly extend the total demonstration length to above 32K. Therefore, this task becomes a natural testbed for long-context understanding.

\begin{figure}[t]
\centering
\includegraphics[width=0.9\textwidth]
{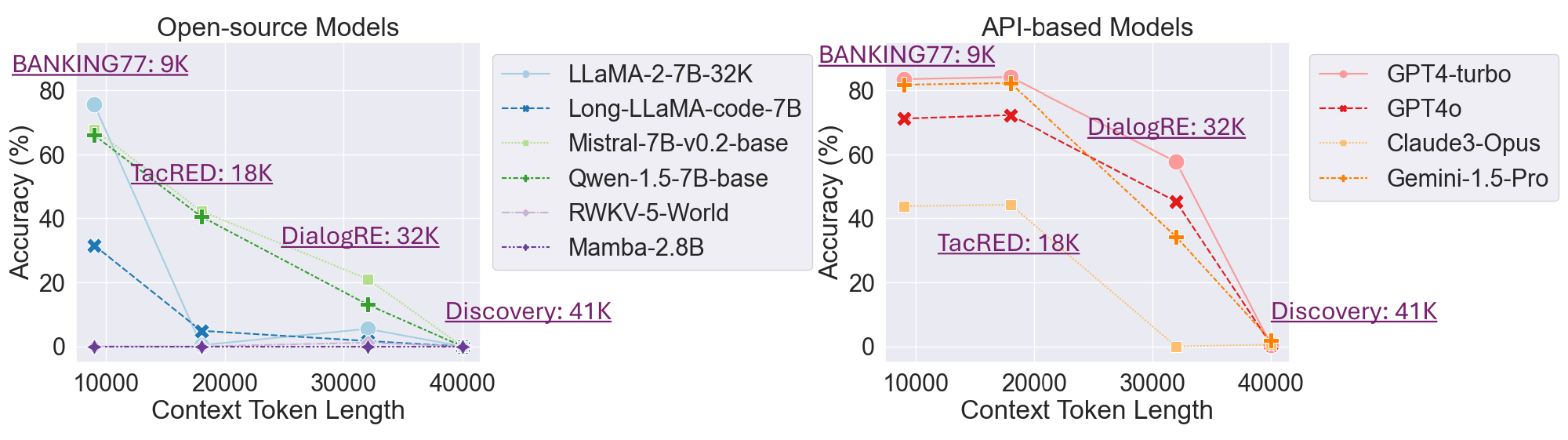}
\vspace{-2ex}
  \caption{Results for representative models across different evaluation datasets. The performance greatly decreases as the task becomes more challenging.}
\vspace{-3ex}
\label{fig:LICL_curve}
\end{figure}

To systematically assess how these extended input capabilities affect model performance in the realm of fine-grained text classification with in-context learning, we have compiled a benchmark, i.e. \eval, consisting of six carefully-selected tasks with different difficulty levels in terms of context length and label space. We evaluate the performance of a wide range of long-context LLMs and \textbf{find that the performance of the open-source models uniformly dips as the task becomes more complex (e.g. requiring longer demonstration) as shown in Figure~\ref{fig:LICL_curve}. Among the open-source models, the non-Transformer-based models, like RWKV and Mamba~\citep{peng2023rwkv, gu2023mamba}, perform far behind the Transformer-based models. Simultaneously, within a task, most of the models can benefit from the extensive demonstration if the length is within a certain range. As the input grows longer, it either hurts or makes the performance fluctuate as shown in Figure~\ref{fig:demo}}. 

On the most difficult extreme-label classification task Discovery~\citep{sileo-etal-2019-mining}, all LLMs achieve close-to-zero performance except Gemini-1.5-Pro with 14\% accuracy. In contrast, a fine-tuned BERT model~\citep{kenton2019bert} can achieve 87\%. This highlights the challenges that the long in-context learning pose for the existing LLMs. Moreover, we make further analysis on the distribution of label position to investigate the factors that affect the long in-context learning capability of these models. It is shown that the position distribution of instances in the prompt can dramatically influence the performance of some of the evaluated models. 

In a nutshell, our contributions to this work can be summarized as follows:
\begin{enumerate}[leftmargin=*, itemsep=0ex]
    \item We have identified in-context learning on extreme-label classification tasks as an ideal testbed for the evaluation of the long-context capability of the current LLMs. We developed \eval, which serves as a complement to earlier benchmarks that concentrated on tasks like long document summarization, question answering (QA), or retrieval, focusing instead on long in-context learning.
    \item We evaluate a line of recent long-context LLMs on \eval and reveal their performances with gradually changed difficulty levels. Simultaneously, we find the sensitivity of some of the long-context LLMs regarding instance position in the prompt. We hope the evaluation results can provide more insights for the improvement of the design of long-context large language models.
\end{enumerate}


\section{Related Work}

\textbf{Long In-context Learning on LLMs} As pre-trained language models continue to grow in size, in-context learning (ICL) has emerged as a favored approach for addressing a wide array of tasks without the need for extensive fine-tuning~\citep{dong2023survey}. A body of research has established that increasing the number of examples demonstrations can enhance ICL performance~\citep{liu-etal-2022-makes, wu2023selfadaptive}. Nonetheless, there are studies indicating that longer input prompts can actually diminish performance~\citep{Liu2023LostIT}, with the effectiveness of prior large language models (LLMs) being constrained by the maximum sequence length encountered during their training. It is also claimed in previous works that LLM+ICL falls short on specification-heavy tasks due to inadequate long-text understanding ability~\citep{peng2023does}. To counter this issue, various works have introduced memory augmentation and extrapolation techniques to support ICL with an extensive set of demonstrations~\citep{li2023incontext, wang2023augmenting}. 

\textbf{Long Context Techniques over LLMs} 
The effectiveness of Transformer-based models is hindered by the quadratic increase in computational cost relative to sequence length, particularly in handling long context inputs. Recent efforts have explored various strategies to address this challenge. Some studies have pursued continued fine-tuning of the LLM with longer context inputs~\citep{rozière2024code, tworkowski2023focused}. Others have leveraged position extrapolation or interpolation, building upon relative rotary positional embedding~\citep{Su2021RoFormerET}, to extend input length beyond the training phase~\citep{alibi, Chen2023ExtendingCW}. Additionally, more approaches have been proposed to mitigate computational issues, including sliding memory window and chunk segmentation~\citep{Hao2022StructuredPS, ratner-etal-2023-parallel, zhu2024pose}. Furthermore, alternative architectures beyond Transformer have been explored to handle long inputs more naturally, such as selective-state-spaces models~\citep{peng2023rwkv, gu2023mamba}. These diverse approaches claim that they can enhance the capabilities of LLMs in processing long context inputs more efficiently.

\textbf{Long Context Evaluation} 
Due to the imperious demands for the support of long-range LLMs, there is a series of benchmarks focusing on long context evaluation. Long-Range Arena~\citep{tay2021long} includes tasks consisting of sequences ranging from 1K to 16K tokens to evaluate variations of fast Transformers. 
LongBench~\citep{bai2023longbench} comprises 21 bilingual datasets with an average length of around 6k words, which have been processed in a unified format to enable effortless evaluation. L-Eval Benchmark~\citep{an2023leval} supports 20 sub-tasks with input lengths of 3K to 200K tokens. LooGLE~\citep{li2023loogle} focuses on summarization and long dependency QA tasks with test instances exceeding 100k words. Most recently, $\infty$Bench~\citep{zhang2024inftybench} encompasses 12 tasks with an average length of 200K tokens. Another recent work explores the impact of extending input lengths, especially on reasoning tasks~\citep{levy2024task}. 

\textbf{Extreme-label Classification}
Extreme-label Classification involves categorizing data into one of an extremely large number of labels, and finds application across a variety of real-world domains such as emotion classification, named entity recognition, and biological function prediction, each requiring precise differentiation among vast label spaces~\citep{zhang2017tacred, sileo-etal-2019-mining, demszky-etal-2020-goemotions, ding-etal-2021-nerd}. Previous methods to tackle Extreme-label Classification tasks range from embedding-based approaches to fine-tuned retrievals~\citep{Bhatia2015SparseLE, vulic-etal-2021-convfit}.
However, integrating this task with long-context large language models presents unique challenges. The large scale of the label space complicates the in-context learning process, where LLMs are expected to discern fine-grained differences among labels based on extensive context~\citep{milios2023incontext}. These challenges make the proposed \eval with a range of difficulty levels a good testing scenario to evaluate the capability of long-context large language models.

\section{Long In-context Evaluation}

\begin{table}[t]
\small
\centering
\caption{Statistics of the collected sub-dataset in \eval. We evaluate from 1-shot/label to 5-shot/label, which results in the shown \#total token range.}
\begin{tabular}{lcccc}
\toprule
\multicolumn{1}{l}{\textbf{Dataset}}  &\multicolumn{1}{c}{\textbf{Task Type}}  &\multicolumn{1}{c}{\textbf{\# Classes}}  & \multicolumn{1}{c}{\textbf{\# Tokens/Shot}}  & \multicolumn{1}{c}{\textbf{\# Total Tokens}} \\
\midrule

\multicolumn{1}{l}{GoEmotion} & Emotion Classification   & 28   &  28  & [1K, 4K]  \\

\multicolumn{1}{l}{BANKING77} & Intent Classification    & 77   & 28   &    [2K, 11K] \\

\multicolumn{1}{l}{TacRED}    & Relation Extraction      & 41   &  80  & [4K, 18K]  \\

\multicolumn{1}{l}{Few-NERD}  & Entity Recognition       & 66   &  61  &  [5K, 23K]    \\

\multicolumn{1}{l}{DialogRE} & Relation Extraction       &  36  & 226   &    [8K, 32K] \\

\multicolumn{1}{l}{Discovery} & Discourse Marker Classification & 174  &  61  &   [10K, 50K] \\
\bottomrule
\end{tabular}
\label{table:benchmark}
\end{table}

\subsection{Long In-context Benchmark}
\label{sec:data}
To support the evaluation of long in-context learning on extreme-label classification tasks in different domains and various difficulty levels, we collect six datasets containing context length from short to long. In order to balance the sequence token length within each dataset and the goal of evaluation for long in-context learning, we keep a subset of the classes among all the classes to format evaluation sets around 1 round, 2 rounds, 3 rounds, 4 rounds, and 5 rounds correspondingly, where each round represent a complete set of examples containing all unique chosen labels. We sample the number of instances from each of the classes evenly to reduce the bias resulting from the label distribution. The statistics of the datasets are described in detail in Table~\ref{table:benchmark}.

\textbf{GoEmotions}~\citep{demszky-etal-2020-goemotions} is the largest manually annotated dataset of 58k English comments from Reddit, which is labeled into 27 emotion categories or Neutral. Each selected example contains 28 tokens on average. 

\textbf{BANKING77}~\citep{casanueva-etal-2020-efficient} is a banking-domain intent detection dataset comprising 13,083 annotated examples over 77 intents. We keep all of the types of intents, and each of the instances contains around 28 tokens.

\textbf{TacRED}~\citep{zhang2017tacred} is a large-scale relation extraction dataset with 106,264 examples built over news and web text. Only one relation is labeled for each of the sentences in the dataset. It covers 41 relation types in total, with an average length of 80 tokens for each example.

\textbf{Few-NERD}~\citep{ding-etal-2021-nerd} is a human-annotated name entity recognition dataset with a hierarchy of 8 coarse-grained and 66 fine-grained entity types. Each of the instances is a paragraph with about 61 tokens on average and contains one or multiple entity names as the ground truth answer. 

\textbf{DialogRE}~\citep{yu-etal-2020-dialogue} is a human-annotated dialogue-based relation extraction dataset from an American television comedy, Friends, with 36 possible relation types existing between an argument pair in a dialogue. Each example contains an average of 226 tokens.

\textbf{Discovery}~\citep{sileo-etal-2019-mining} automatically discovers sentence pairs with relevant discourse markers and forms a dataset containing 174 discourse markers with at least 10K examples each. Each example contains around 61 tokens. This dataset is the most difficult task with fine-grained labels.

\subsection{Model and Experimental Setup}
\label{model}

In the exploration of in-context learning for extreme-label classification, we conduct a comprehensive evaluation of popular open-source long-context language models of size around 7B parameters. We also include SoTA models like Gemini-1.5-Pro, Claude3-Opus, and GPT-4-turbo. \autoref{tab:models} provides an overview of the models investigated, highlighting the innovations in their architecture specifically for dealing with long context. We can observe that there are multiple strategies adopted to extend the context window. Some of the models support the training context window size while some models support length extrapolation. RWKV~\citep{peng2023rwkv} and Mamba~\citep{gu2023mamba} are the two new RNN-like architectures to decrease attention complexity, which would allow the model to easily extrapolate to much longer inputs with linear time/memory complexity.
\begin{table}[!tbh]
\small
\caption{The overview of the evaluated models. We utilize base models before instruction-tuning except API-based models. LF means fine-tuning the model on longer-context corpus after pre-training. }
\begin{tabular}{llllll}
\toprule
Model                & Size & Initialization & Strategy                 & Train & Support \\
\midrule
\defcitealias{team2024gemma}{Gemma-7B-base}\citetalias{team2024gemma}        
& 7B   & Gemma          &  RoPE + LF    & 8K    & 8K      \\
\defcitealias{Liu2023LostIT}{LLaMA-2-7B-32K}\citetalias{Liu2023LostIT}    
& 7B   & LLaMA-2        & Position Interpolation   & 32K   & 32K     \\
\defcitealias{zeng2022glm}{ChatGLM3-6B-32K}\citetalias{zeng2022glm}
& 6B   & ChatGLM        & Position Encoding Scheme & 32K   & 32K     \\
\defcitealias{bai2023qwen}{Qwen-1.5-7B-base}\citetalias{bai2023qwen}
& 7B   & Qwen           & NTK-Aware Interpolation  & 32K   & 32K     \\
\defcitealias{jiang2023mistral}{Mistral-7B-v0.2-base}\citetalias{jiang2023mistral}
& 7B   & Mistral        & LF                       & 32K   & 32K     \\
\defcitealias{Chen2023LongLoRAEF}{LLaMA-2-7B-LongLora}\citetalias{Chen2023LongLoRAEF}
& 7B   & LLaMA-2        & Shifted Short Attention  & 100K  & 100K     \\
\defcitealias{ai2024yi}{Yi-6B-200K}\citetalias{ai2024yi}
& 6B   & Yi             & Position Interpolation +LF   & 200K  & 200K    \\
\defcitealias{cai2024intern}{InternLM2-7B-base}\citetalias{cai2024intern}
& 7B   & InternLM       & Dynamic NTK              & 32K   & 200K   \\
\defcitealias{tworkowski2023focused}{Long-LLaMA-code-7B}\citetalias{tworkowski2023focused}
& 7B   & LLaMA-2        & Focused Transformer      & 8K    & 256K    \\
\midrule
\defcitealias{peng2023rwkv}{RWKV-5-World}\citetalias{peng2023rwkv}
& 3B   & RWKV           & Attention-free Model     & 4K    & $\infty$       \\
\defcitealias{gu2023mamba}{Mamba-2.8B}\citetalias{gu2023mamba}
& 2.8B & Mamba          & State Space Model        & 2K    & $\infty$       \\
\midrule
\defcitealias{achiam2023gpt}{GPT4-turbo}\citetalias{achiam2023gpt}
& -    & GPT-4          & -                        & -     & 128K    \\
\defcitealias{achiam2023gpt}{GPT4o}\citetalias{achiam2023gpt}
& -    & GPT-4          & -                        & -     & 128K    \\
\defcitealias{TheC3}{Claude3-Opus}\citetalias{TheC3}
& -    & Claude3        & -                        & -  & 200K \\
\defcitealias{Reid2024Gemini1U}{Gemini-1.5-Pro}\citetalias{Reid2024Gemini1U}
& -    & Gemini         & -           & -   & 10M     \\
\bottomrule
\end{tabular}
\vspace{-3ex}
\label{tab:models}
\end{table}

We construct a prompt following the template as shown in \ref{sec:prompt} for each of the datasets. To fairly evaluate the open-source and API-based models with a series of input lengths, we sample the same example set for all the models with labels distributed evenly to ensure an unbiased distribution for the in-context demonstration. For instance, an input of one round will include one set of examples traversing all the types, and 5 rounds will contain instances from each of the labels 5 times. For testing, we sample 500 examples from the test set of each dataset, simultaneously ensuring an even distribution in terms of the type of labels. All the open-source models are loaded from the weights in HuggingFace\footnote{\url{https://huggingface.co}}, and inferred on eight NVIDIA RTX A6000 GPUs, while the API-based models are based on the official documentations \footnote{\url{https://platform.openai.com/docs/guides/text-generation/chat-completions-api}, \url{https://cloud.google.com/vertex-ai/generative-ai/docs/multimodal/overview}}.

\subsection{Experiment Result}

The main evaluation results are demonstrated in Table~\ref{tab:bank77}, Table~\ref{tab:tacred}, Table~\ref{tab:dial}, Table~\ref{tab:disc} and \autoref{sec:add_data}. For the entity recognition and relationship extraction dataset, we use the F1 score as the evaluation metric, and Accuracy is utilized for the other datasets. From the presented results, generally, we can find that models of Transformer-based architecture perform consistently better than the RNN-based ones in all the evaluated datasets. However, both of them are still falling behind the powerful API-based models. For a relatively simple task like BANKING77, whose context length from 1 round to 5 rounds is 2K to 14 K, most of the models can benefit from the extensive context with more demonstrations. As shown in Figure~\ref{fig:demo} and Table~\ref{tab:bank77}, from 2K to 4K, there is either a huge increase nearly doubling the accuracy, or a complete failure for most of the open-source models. After 3 rounds, limited performance gain can be achieved by adding more examples. 
When it comes to more complicated tasks like TacRED and DialogueRE in Table~\ref{tab:tacred} and Table~\ref{tab:dial}, which are more urgently requiring the capability of long-context comprehension, the overall performance of all the few-shot models drops compared to BANKING77. As shown in the middle plot of Figure~\ref{fig:demo}, only GPT4-turbo and GPT4o can consistently benefit from more demonstrations, all of the other models reach their peak at the middle with context length around 13K to 25K.

For the most challenging Discovery dataset, which has an extremely large label space including 174 classes, one round of traversing for all the label possibilities has already made up a context length of 10K. In this extreme case, all of the models except Gemini-1.5-Pro, fail to tell the difference among the fine-grained types including GPT4-turbo, leading to a score of 0. The results across different datasets reveal the models' capability to understand different types of tasks. Our initial hypothesis suggests that the strongest LLMs like GPT-4-turbo are capped at a certain complexity level between DialogRE and Discovery. 

Another interesting observation we have is that some LLMs' performance on the extreme-label ICL seems highly predictable. According to the left sub-graph of ~\autoref{fig:LICL_curve}, the performance of Qwen and Mistral is almost linear w.r.t the demonstration length. This reveals that there might be an underlying mathematical relation between performance and the task complexity for ICL. 


\begin{table}[tbh]
\small
\caption{BANKING77 result with respect to increasing context length. \textbf{1R} represents one round of traversing all the instances with a unique label.}
\centering
\begin{tabular}{lccccccc}
\toprule
\multicolumn{1}{l}{\textbf{Model}} &\multicolumn{1}{c}{\textbf{Param}}
&\multicolumn{1}{c}{\textbf{Support}}  &\multicolumn{1}{c}{\textbf{1R}} &\multicolumn{1}{c}{\textbf{2R}}  &\multicolumn{1}{c}{\textbf{3R}}    &\multicolumn{1}{c}{\textbf{4R}}  &\multicolumn{1}{c}{\textbf{5R}}               \\
\midrule
\multicolumn{1}{l}{\textbf{Context Tokens}} &\multicolumn{1}{c}{\textbf{}}
&\multicolumn{1}{c}{\textbf{}}  &\multicolumn{1}{c}{\textbf{2K}} &\multicolumn{1}{c}{\textbf{4K}}  &\multicolumn{1}{c}{\textbf{7K}}    &\multicolumn{1}{c}{\textbf{9K}}  &\multicolumn{1}{c}{\textbf{14K}}  \\
\midrule

\multicolumn{1}{l}{Gemma-7B-base} & 7B & 8K & 0  & 0	& 0	& 0	& 0 \\

\multicolumn{1}{l}{LLaMA-2-7B-32K} & 7B & 32K & \textbf{30.2}  & \textbf{70.4} & \textbf{72.0}	& \textbf{75.6}	& \textbf{77.2}	\\

\multicolumn{1}{l}{ChatGLM3-6B-32K} & 6B & 32K & 16.6  & 23.2	& 22.4	& 22.8	& 8.8  \\

\multicolumn{1}{l}{Qwen-1.5-7B-base} & 7B & 32K & 21.6  & 52.8	& 61.4	& 66.0	& 67.8 \\

\multicolumn{1}{l}{Mistral-7B-v0.2-base} & 7B & 32K & 29.8  & 43.6	& 66.4	& 67.8	& 64.0 \\

\multicolumn{1}{l}{LLaMA-2-7B-LongLora} & 7B & 100K & 0  & 0	& 0	& 0	& 0 \\

\multicolumn{1}{l}{Yi-6B-200K} & 6B & 200K & 25.8  & 0	& 0	& 0	& 1.2 \\


\multicolumn{1}{l}{InternLM2-7B-base} & 7B & 200K & 5.6  & 0	& 0	& 0	& 0 \\

\multicolumn{1}{l}{Long-LLaMA-code-7B} & 7B & 256K & 3.0  & 19.4	& 28.0	& 31.6	& 32.6  \\

\midrule

\multicolumn{1}{l}{RWKV-5-World} & 7B & 4K & 8.6  & 21.2	& 0.4	& 0	& 0 \\

\multicolumn{1}{l}{Mamba-2.8B} & 2.8B & 2K & 0  & 0	& 0	& 0	& 0 \\

\midrule


\multicolumn{1}{l}{GPT4-turbo} & N/A & 128K & \textbf{73.5}  & \textbf{80.5}	& 82.0	&  \textbf{83.5}	&  \textbf{84.4} \\

\multicolumn{1}{l}{GPT4o} & N/A & 128K & 80.8  & 79.8	& 81.2	&  71.2 & 71.4 \\

\multicolumn{1}{l}{Claude3-Opus} & N/A & 200K & 60.0  & 62.6	& 62.2	&  43.8	&  26.0 \\

\multicolumn{1}{l}{Gemini-1.5-Pro} & N/A & 10M & 28.8  & 79.4	& \textbf{82.2}	&  81.8	&  70.4 \\

\midrule
\multicolumn{1}{l}{SoTA (RoBERTA + ICDA)} & N/A & - & \multicolumn{5}{c}{\textbf{94.4}} \\

\bottomrule
\end{tabular}
\label{tab:bank77}
\vspace{2ex}
\caption{TacRED result with respect to increasing context length.}
\begin{tabular}{lccccccc}

\toprule
\multicolumn{1}{l}{\textbf{Model}} &\multicolumn{1}{c}{\textbf{Param}}
&\multicolumn{1}{c}{\textbf{Support}}  &\multicolumn{1}{c}{\textbf{1R}} &\multicolumn{1}{c}{\textbf{2R}}  &\multicolumn{1}{c}{\textbf{3R}}    &\multicolumn{1}{c}{\textbf{4R}}  &\multicolumn{1}{c}{\textbf{5R}}               \\
\midrule
\multicolumn{1}{l}{\textbf{Context Tokens}} &\multicolumn{1}{c}{\textbf{}}
&\multicolumn{1}{c}{\textbf{}}  &\multicolumn{1}{c}{\textbf{4K}} &\multicolumn{1}{c}{\textbf{7K}}  &\multicolumn{1}{c}{\textbf{10K}}    &\multicolumn{1}{c}{\textbf{14K}}  &\multicolumn{1}{c}{\textbf{18K}}               \\
\midrule

\multicolumn{1}{l}{Gemma-7B-base} & 7B & 8K & 0.4  & 0.4	& 0	& 0	& 0 \\

\multicolumn{1}{l}{LLaMA-2-7B-32K} & 7B & 32K & 0  & 0.4	& 0.4	& 0.8	& 0.4	\\

\multicolumn{1}{l}{ChatGLM3-6B-32K} & 6B & 32K & 29.7  & 36.1	& 38.9	&  40.1 & 25.2 \\

\multicolumn{1}{l}{Qwen-1.5-7B-base} & 7B & 32K & 38.7  &	47.3 & 45.2 & 43.6	& 40.6 \\

\multicolumn{1}{l}{Mistral-7B-v0.2-base} & 7B & 32K & \textbf{53.3}  & \textbf{53.1}	& \textbf{51.6}	& \textbf{48.0}	& \textbf{42.3} \\

\multicolumn{1}{l}{LLaMA-2-7B-LongLora} & 7B & 100K & 0  & 0	& 0	& 0	& 0 \\

\multicolumn{1}{l}{Yi-6B-200K} & 6B & 200K & 5.6  & 1.9	& 8.0	& 9.5	& 2.0 \\


\multicolumn{1}{l}{InternLM2-7B-base} & 7B & 200K & 29.6  & 27.2	& 15.5	& 10.7	& 8.0 \\

\multicolumn{1}{l}{Long-LLaMA-code-7B} & 7B & 256K & 3.8  & 7.1	& 4.1	& 6.6	& 4.9  \\

\midrule

\multicolumn{1}{l}{RWKV-5-World} & 7B & 1K & 2.3  & 2.6	& 1.0	& 0	& 1.2 \\

\multicolumn{1}{l}{Mamba-2.8B} & 2.8B & 2K & 0  & 0	& 0	& 0	& 0 \\

\midrule


\multicolumn{1}{l}{GPT4-turbo} & N/A & 128K   & \textbf{74.4}  & 76.5	& 79.5	& 80.4	&  \textbf{84.2} \\

\multicolumn{1}{l}{GPT4o} & N/A & 128K   & 71.1  & 75.5	& 73.6	& 73.2	&  72.3 \\

\multicolumn{1}{l}{Claude3-Opus} & N/A & 200K & 68.7  & 74.1	& 35.4	& 43.4	& 44.3 \\

\multicolumn{1}{l}{Gemini-1.5-Pro} & N/A & 10M & 72.6  & \textbf{81.4}	& \textbf{79.6}	&  \textbf{81.4}	&  82.3 \\

\midrule
\multicolumn{1}{l}{SoTA (DeepStruct)} & N/A & - & \multicolumn{5}{c}{76.8} \\

\bottomrule
\end{tabular}
\label{tab:tacred}

\end{table}
\begin{table}[tbh]
\small
\centering
\caption{DialogRE result with respect to increasing context length.}
\begin{tabular}{lccccccc}

\toprule
\multicolumn{1}{l}{\textbf{Model}} &\multicolumn{1}{c}{\textbf{Param}}
&\multicolumn{1}{c}{\textbf{Support}}  &\multicolumn{1}{c}{\textbf{1R}} &\multicolumn{1}{c}{\textbf{2R}}  &\multicolumn{1}{c}{\textbf{3R}}    &\multicolumn{1}{c}{\textbf{4R}}  &\multicolumn{1}{c}{\textbf{5R}}               \\
\midrule
\multicolumn{1}{l}{\textbf{Context Tokens}} &\multicolumn{1}{c}{\textbf{}}
&\multicolumn{1}{c}{\textbf{}}  &\multicolumn{1}{c}{\textbf{8K}} &\multicolumn{1}{c}{\textbf{13K}}  &\multicolumn{1}{c}{\textbf{19K}}    &\multicolumn{1}{c}{\textbf{25K}}  &\multicolumn{1}{c}{\textbf{32K}}               \\
\midrule

\multicolumn{1}{l}{Gemma-7B-base} & 7B & 8K & 14.7 & 0	& 0	& 0	& 0 \\

\multicolumn{1}{l}{LLaMA-2-7B-32K} & 7B & 32K & 6.6 & 13.5	& 6.0	& 5.4	& 5.5\\

\multicolumn{1}{l}{ChatGLM3-6B-32K} & 6B & 32K & 0.5 & 1.1	& 2.5	& 1.8	& 7.6  \\

\multicolumn{1}{l}{Qwen-1.5-7B-base} & 7B & 32K & 14.0 & 17.8	& 15.3	& 16.2	& 13.1 \\

\multicolumn{1}{l}{Mistral-7B-v0.2-base} & 7B & 32K & \textbf{24.0} & \textbf{23.0}	& \textbf{23.2}	& \textbf{22.0}	& \textbf{21.1} \\

\multicolumn{1}{l}{LLaMA-2-7B-LongLora} & 7B & 100K & 0 & 0	& 0	& 0	& 0 \\

\multicolumn{1}{l}{Yi-6B-200K} & 6B & 200K & 0 & 0	& 0.4	& 0.4	& 0 \\



\multicolumn{1}{l}{InternLM2-7B-base} & 7B & 200K & 12.0 & 13.2 & 5.8 & 1.8 & 0.7 \\

\multicolumn{1}{l}{Long-LLaMA-code-7B} & 7B & 256K & 2.7 & 3.0	& 2.6	& 5.2	& 1.7 \\

\midrule

\multicolumn{1}{l}{RWKV-5-World} & 7B & 4K & 0 & 0	& 0	& 0	& 0 \\

\multicolumn{1}{l}{Mamba-2.8B} & 2.8B & 2K & 0 & 0	& 0	& 0	& 0 \\

\midrule


\multicolumn{1}{l}{GPT4-turbo} & N/A & 128K & \textbf{42.9} & \textbf{47.8} & \textbf{52.0} & \textbf{55.9} & \textbf{57.7} \\

\multicolumn{1}{l}{GPT4o} & N/A & 128K & 40.6 & 41.5 & 41.0 & 47.3 & 45.3 \\

\multicolumn{1}{l}{Claude3-Opus} & N/A & 200K & 16.8 & 30.3	& 15.3	& 0.8	& 0 \\

\multicolumn{1}{l}{Gemini-1.5-Pro} & N/A & 10M & 29.6  & 37.8	& 31.2	&  32.4	&  34.3 \\

\midrule

\multicolumn{1}{l}{SoTA (HiDialog)} & N/A & - & \multicolumn{5}{c}{\textbf{77.1}} \\

\bottomrule
\end{tabular}
\label{tab:dial}
\vspace{2ex}

\caption{Discovery result with respect to increasing context length.}
\begin{tabular}{lccccccc}

\toprule
\multicolumn{1}{l}{\textbf{Model}} &\multicolumn{1}{c}{\textbf{Param}}
&\multicolumn{1}{c}{\textbf{Support}}  &\multicolumn{1}{c}{\textbf{1R}} &\multicolumn{1}{c}{\textbf{2R}}  &\multicolumn{1}{c}{\textbf{3R}}    &\multicolumn{1}{c}{\textbf{4R}}  &\multicolumn{1}{c}{\textbf{5R}}               \\
\midrule
\multicolumn{1}{l}{\textbf{Context Tokens}} &\multicolumn{1}{c}{\textbf{}}
&\multicolumn{1}{c}{\textbf{}}  &\multicolumn{1}{c}{\textbf{10K}} &\multicolumn{1}{c}{\textbf{20K}}  &\multicolumn{1}{c}{\textbf{30K}}    &\multicolumn{1}{c}{\textbf{40K}}  &\multicolumn{1}{c}{\textbf{50K}}  \\
\midrule

\multicolumn{1}{l}{Gemma-7B-base} & 7B & 8K & 0  & 0	& 0	& 0	& 0 \\

\multicolumn{1}{l}{LLaMA-2-7B-32K} & 7B & 32K & 0  & 0	& 0	& 0	& \xmark	\\

\multicolumn{1}{l}{ChatGLM3-6B-32K} & 6B & 32k & 0  & \textbf{1.0}	& 0	& \xmark	& \xmark  \\

\multicolumn{1}{l}{Qwen-1.5-7B-base} & 7B & 32K & 0  & 0	& 0	& 0	& 0 \\

\multicolumn{1}{l}{Mistral-7B-v0.2-base} & 7B & 32K & 0  & 0	& 0	& 0	& 0 \\

\multicolumn{1}{l}{LLaMA-2-7B-LongLora} & 7B & 100K & 0  & 0	& 0	& 0	& 0 \\

\multicolumn{1}{l}{Yi-6B-200K} & 6B & 200k & 0  & 0 & 0	& 0	& 0 \\


\multicolumn{1}{l}{InternLM2-7B-base} & 7B & 200K & 0  & 0	& 0	& 0	& 0 \\

\multicolumn{1}{l}{Long-LLaMA-code-7B} & 7B & 256K & 0  & 0	& 0	& 0	& 0  \\

\midrule

\multicolumn{1}{l}{RWKV-5-World} & 7B & 4K & 0  & 0.2	& 0	& 0	& 0 \\

\multicolumn{1}{l}{Mamba-2.8B} & 2.8B & 2K & 0  & 0	& 0	& 0	& 0 \\

\midrule


\multicolumn{1}{l}{GPT4-turbo} & N/A & 128K & 1.5  & 0.5	& 0.5	& 0.5	& 0.5 \\

\multicolumn{1}{l}{GPT4o} & N/A & 128K & 2.8  & 0.8	& 0.8	&  0.6	&  0.4 \\

\multicolumn{1}{l}{Claude3-Opus} & N/A & 200K & 1.2  & 0.6	& 0.6	& 0.6 &  0.2 \\

\multicolumn{1}{l}{Gemini-1.5-Pro} & N/A & 10M & \textbf{14.0}  & \textbf{6.0}	& \textbf{3.2}	&  \textbf{1.8}	&  \textbf{2.8} \\

\midrule

\multicolumn{1}{l}{SoTA (MTL)} & N/A & - & \multicolumn{5}{c}{\textbf{87.4}} \\

\bottomrule
\end{tabular}

\label{tab:disc}

\end{table}

\section{Exploratory Experiment}
Inspired by the Lost in the Middle phenomenon~\citep{Liu2023LostIT}, we take analysis experiments to explore whether the position distribution of the instances will make a difference in the performance for long in-context learning with extreme-label classification tasks.

\subsection{Scattered Distribution}
\begin{figure}[t!]
\includegraphics[width=\textwidth]
{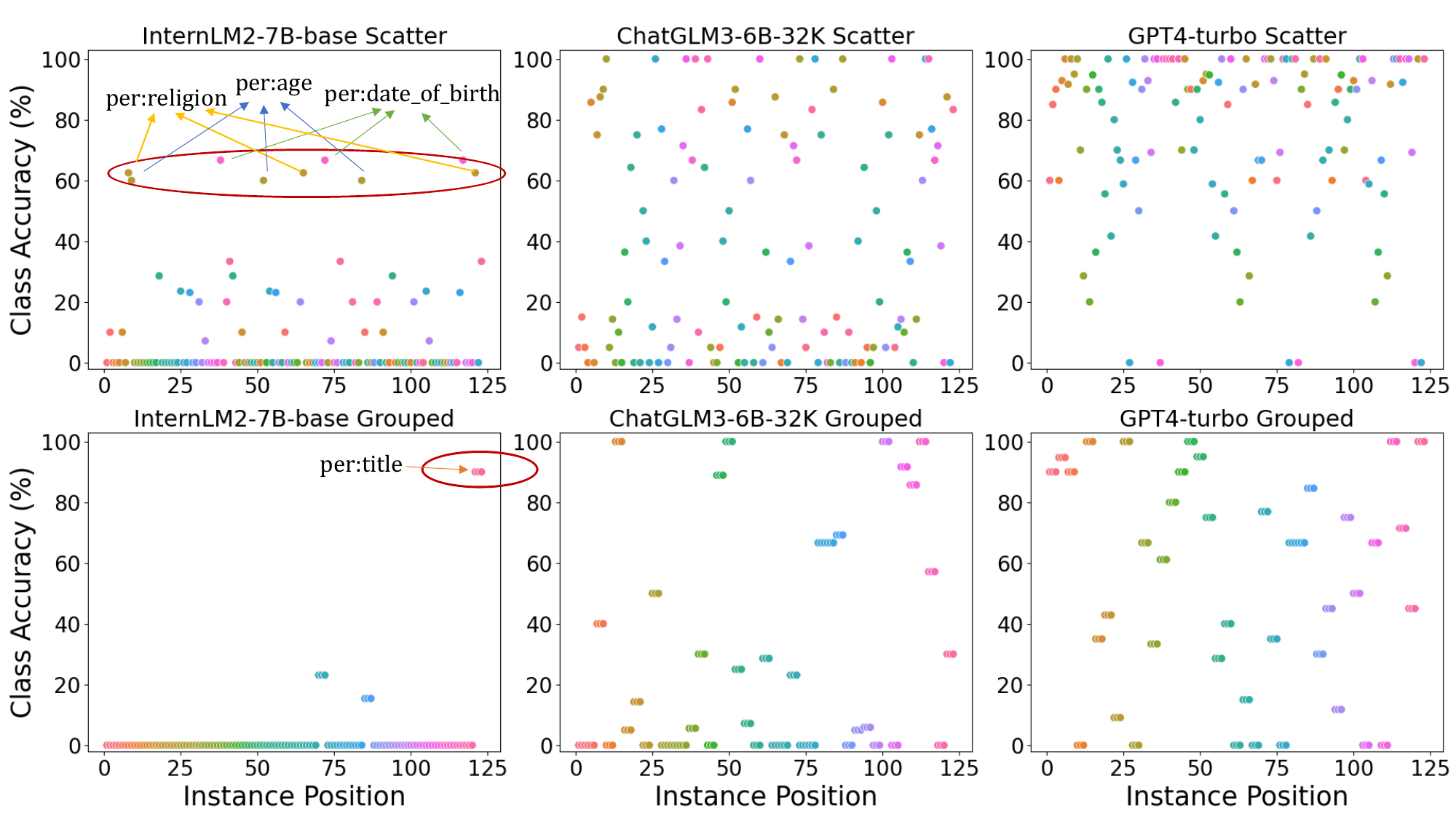}
  \caption{Visualization of accuracy for every class when instances from the same class are scattered V.S. grouped in the demonstration prompt.}
\label{fig:group_ana}
\vspace{-3ex}
\end{figure}

In our investigation, we conduct pilot experiments on TacRED, a medium-complexity dataset, with each label type demonstrated three times, resulting in a total of 123 distinct instances (calculated as $41 \times 3$). Within these experiments, instances bearing the same labels are distributed randomly to form a scattered configuration. For each instance, we track its relative position within the prompt alongside its corresponding label, thereafter computing the accuracy for each label class. As illustrated in the first row of Figure~\ref{fig:group_ana}, the visualization delineates the accuracy of each label, aligned with its position within the prompt, where diverse colors symbolize various label types. In scenarios where class instances are scattered, certain models, such as InternLM2-7B-base, demonstrate acceptable performances—approximately 60\% accuracy merely on specific labels, as highlighted by a red circle in Figure~\ref{fig:group_ana}, regardless of the instance placements. Conversely, other models, like ChatGLM3-6B-32K, exhibit robust performance across a broad spectrum of labels. Remarkably, the GPT4-turbo model consistently surpasses an 80\% accuracy threshold for the majority of label types, with only a minimal count of exceptions.

\subsection{Grouped Distribution}
To facilitate a clear comparison between scattered and grouped distributions, we organize instances of the same class to be adjacent within the demonstration prompts. The impact of this reorganization on model performance, both pre and post-grouping, is presented in~\autoref{sec:group}. It is easy to observe that there is a general decline in performance across most models after grouping instances by class. Notably, models such as Mistral and InternLM2 exhibit significant performance drops, underscoring a pronounced sensitivity to instance grouping. In an effort to delve deeper into this phenomenon, we visualize the accuracy of grouped labels in relation to their positions within the prompt, as illustrated in Figure~\ref{fig:group_ana}. This visualization reveals that instances of the same class, denoted by dots of the same color, are positioned nearby. It became evident that some models, like InternLM2 or Mistral shown in~\autoref{sec:group}, demonstrate high sensitivity to the distribution of instances, only handling instances with labels positioned at the end of the prompt. Conversely, other open-source models such as ChatGLM3-6B-32K, with a modest 3.3\% drop in accuracy, proved to be more resilient to changes in instance positioning. Surprisingly, even the GPT4-turbo and Gemini1.5-Pro are not immune to the challenges posed by grouped distributions, experiencing a notable decline in performance by 20.3\% and 22.3\%. This observed decrease in performance is consistent across models, unaffected by the specific positions of the labels within the prompt.

\begingroup

\section{Conclusion}
In summary, our research explores the capability of LLMs on long in-context learning tasks, particularly in extreme-label classification scenarios. We curate a dataset \eval consisting of long in-context learning tasks with different difficulty levels in terms of context length. Through our study, we have discovered that LLMs demonstrate dramatic performance degradation when it comes to more difficult tasks. 
Our exploratory experiments further highlight the impact of the distribution of examples within prompts on model performance. We hope  \eval and our findings contribute to the ongoing efforts to enhance LLMs' understanding of long contexts. 


\newpage

\bibliography{neurips_data_2024}

\begin{thebibliography}{58}
\providecommand{\natexlab}[1]{#1}
\providecommand{\url}[1]{\texttt{#1}}
\expandafter\ifx\csname urlstyle\endcsname\relax
  \providecommand{\doi}[1]{doi: #1}\else
  \providecommand{\doi}{doi: \begingroup \urlstyle{rm}\Url}\fi

\bibitem[The()]{TheC3}
The claude 3 model family: Opus, sonnet, haiku.
\newblock URL \url{https://api.semanticscholar.org/CorpusID:268232499}.

\bibitem[Achiam et~al.(2023)Achiam, Adler, Agarwal, Ahmad, Akkaya, Aleman, Almeida, Altenschmidt, Altman, Anadkat, et~al.]{achiam2023gpt}
Josh Achiam, Steven Adler, Sandhini Agarwal, Lama Ahmad, Ilge Akkaya, Florencia~Leoni Aleman, Diogo Almeida, Janko Altenschmidt, Sam Altman, Shyamal Anadkat, et~al.
\newblock Gpt-4 technical report.
\newblock \emph{arXiv preprint arXiv:2303.08774}, 2023.

\bibitem[AI et~al.(2024)AI, :, Young, Chen, Li, Huang, Zhang, Zhang, Li, Zhu, Chen, Chang, Yu, Liu, Liu, Yue, Yang, Yang, Yu, Xie, Huang, Hu, Ren, Niu, Nie, Xu, Liu, Wang, Cai, Gu, Liu, and Dai]{ai2024yi}
01. AI, :, Alex Young, Bei Chen, Chao Li, Chengen Huang, Ge~Zhang, Guanwei Zhang, Heng Li, Jiangcheng Zhu, Jianqun Chen, Jing Chang, Kaidong Yu, Peng Liu, Qiang Liu, Shawn Yue, Senbin Yang, Shiming Yang, Tao Yu, Wen Xie, Wenhao Huang, Xiaohui Hu, Xiaoyi Ren, Xinyao Niu, Pengcheng Nie, Yuchi Xu, Yudong Liu, Yue Wang, Yuxuan Cai, Zhenyu Gu, Zhiyuan Liu, and Zonghong Dai.
\newblock Yi: Open foundation models by 01.ai, 2024.

\bibitem[An et~al.(2023)An, Gong, Zhong, Zhao, Li, Zhang, Kong, and Qiu]{an2023leval}
Chenxin An, Shansan Gong, Ming Zhong, Xingjian Zhao, Mukai Li, Jun Zhang, Lingpeng Kong, and Xipeng Qiu.
\newblock L-eval: Instituting standardized evaluation for long context language models, 2023.

\bibitem[Anil et~al.(2022)Anil, Wu, Andreassen, Lewkowycz, Misra, Ramasesh, Slone, Gur-Ari, Dyer, and Neyshabur]{anil2022exploring}
Cem Anil, Yuhuai Wu, Anders~Johan Andreassen, Aitor Lewkowycz, Vedant Misra, Vinay~Venkatesh Ramasesh, Ambrose Slone, Guy Gur-Ari, Ethan Dyer, and Behnam Neyshabur.
\newblock Exploring length generalization in large language models.
\newblock In Alice~H. Oh, Alekh Agarwal, Danielle Belgrave, and Kyunghyun Cho (eds.), \emph{Advances in Neural Information Processing Systems}, 2022.
\newblock URL \url{https://openreview.net/forum?id=zSkYVeX7bC4}.

\bibitem[Bai et~al.(2023{\natexlab{a}})Bai, Bai, Chu, Cui, Dang, Deng, Fan, Ge, Han, Huang, Hui, Ji, Li, Lin, Lin, Liu, Liu, Lu, Lu, Ma, Men, Ren, Ren, Tan, Tan, Tu, Wang, Wang, Wang, Wu, Xu, Xu, Yang, Yang, Yang, Yang, Yao, Yu, Yuan, Yuan, Zhang, Zhang, Zhang, Zhang, Zhou, Zhou, Zhou, and Zhu]{bai2023qwen}
Jinze Bai, Shuai Bai, Yunfei Chu, Zeyu Cui, Kai Dang, Xiaodong Deng, Yang Fan, Wenbin Ge, Yu~Han, Fei Huang, Binyuan Hui, Luo Ji, Mei Li, Junyang Lin, Runji Lin, Dayiheng Liu, Gao Liu, Chengqiang Lu, Keming Lu, Jianxin Ma, Rui Men, Xingzhang Ren, Xuancheng Ren, Chuanqi Tan, Sinan Tan, Jianhong Tu, Peng Wang, Shijie Wang, Wei Wang, Shengguang Wu, Benfeng Xu, Jin Xu, An~Yang, Hao Yang, Jian Yang, Shusheng Yang, Yang Yao, Bowen Yu, Hongyi Yuan, Zheng Yuan, Jianwei Zhang, Xingxuan Zhang, Yichang Zhang, Zhenru Zhang, Chang Zhou, Jingren Zhou, Xiaohuan Zhou, and Tianhang Zhu.
\newblock Qwen technical report, 2023{\natexlab{a}}.

\bibitem[Bai et~al.(2023{\natexlab{b}})Bai, Lv, Zhang, Lyu, Tang, Huang, Du, Liu, Zeng, Hou, Dong, Tang, and Li]{bai2023longbench}
Yushi Bai, Xin Lv, Jiajie Zhang, Hongchang Lyu, Jiankai Tang, Zhidian Huang, Zhengxiao Du, Xiao Liu, Aohan Zeng, Lei Hou, Yuxiao Dong, Jie Tang, and Juanzi Li.
\newblock Longbench: A bilingual, multitask benchmark for long context understanding, 2023{\natexlab{b}}.

\bibitem[Bhatia et~al.(2015)Bhatia, Jain, Kar, Varma, and Jain]{Bhatia2015SparseLE}
Kush Bhatia, Himanshu Jain, Purushottam Kar, Manik Varma, and Prateek Jain.
\newblock Sparse local embeddings for extreme multi-label classification.
\newblock In \emph{Neural Information Processing Systems}, 2015.
\newblock URL \url{https://api.semanticscholar.org/CorpusID:11419932}.

\bibitem[Cai et~al.(2024)Cai, Cao, Chen, ..., Qiao, and Lin]{cai2024intern}
Zheng Cai, Maosong Cao, Haojiong Chen, ..., Yu~Qiao, and Dahua Lin.
\newblock Internlm2 technical report.
\newblock \emph{arXiv preprint arXiv:2403.17297}, 2024.

\bibitem[Casanueva et~al.(2020)Casanueva, Tem{\v{c}}inas, Gerz, Henderson, and Vuli{\'c}]{casanueva-etal-2020-efficient}
I{\~n}igo Casanueva, Tadas Tem{\v{c}}inas, Daniela Gerz, Matthew Henderson, and Ivan Vuli{\'c}.
\newblock Efficient intent detection with dual sentence encoders.
\newblock In Tsung-Hsien Wen, Asli Celikyilmaz, Zhou Yu, Alexandros Papangelis, Mihail Eric, Anuj Kumar, I{\~n}igo Casanueva, and Rushin Shah (eds.), \emph{Proceedings of the 2nd Workshop on Natural Language Processing for Conversational AI}, pp.\  38--45, Online, July 2020. Association for Computational Linguistics.
\newblock \doi{10.18653/v1/2020.nlp4convai-1.5}.
\newblock URL \url{https://aclanthology.org/2020.nlp4convai-1.5}.

\bibitem[Chen et~al.(2023{\natexlab{a}})Chen, Wong, Chen, and Tian]{Chen2023ExtendingCW}
Shouyuan Chen, Sherman Wong, Liangjian Chen, and Yuandong Tian.
\newblock Extending context window of large language models via positional interpolation.
\newblock \emph{ArXiv}, abs/2306.15595, 2023{\natexlab{a}}.
\newblock URL \url{https://api.semanticscholar.org/CorpusID:259262376}.

\bibitem[Chen et~al.(2023{\natexlab{b}})Chen, Qian, Tang, Lai, Liu, Han, and Jia]{Chen2023LongLoRAEF}
Yukang Chen, Shengju Qian, Haotian Tang, Xin Lai, Zhijian Liu, Song Han, and Jiaya Jia.
\newblock Longlora: Efficient fine-tuning of long-context large language models.
\newblock In \emph{The Twelfth International Conference on Learning Representations}, 2023{\natexlab{b}}.

\bibitem[Dasigi et~al.(2021)Dasigi, Lo, Beltagy, Cohan, Smith, and Gardner]{dasigi2021dataset}
Pradeep Dasigi, Kyle Lo, Iz~Beltagy, Arman Cohan, Noah~A Smith, and Matt Gardner.
\newblock A dataset of information-seeking questions and answers anchored in research papers.
\newblock In \emph{Proceedings of the 2021 Conference of the North American Chapter of the Association for Computational Linguistics: Human Language Technologies}, pp.\  4599--4610, 2021.

\bibitem[Demszky et~al.(2020)Demszky, Movshovitz-Attias, Ko, Cowen, Nemade, and Ravi]{demszky-etal-2020-goemotions}
Dorottya Demszky, Dana Movshovitz-Attias, Jeongwoo Ko, Alan Cowen, Gaurav Nemade, and Sujith Ravi.
\newblock {G}o{E}motions: A dataset of fine-grained emotions.
\newblock In Dan Jurafsky, Joyce Chai, Natalie Schluter, and Joel Tetreault (eds.), \emph{Proceedings of the 58th Annual Meeting of the Association for Computational Linguistics}, pp.\  4040--4054, Online, July 2020. Association for Computational Linguistics.
\newblock \doi{10.18653/v1/2020.acl-main.372}.
\newblock URL \url{https://aclanthology.org/2020.acl-main.372}.

\bibitem[Ding et~al.(2021)Ding, Xu, Chen, Wang, Han, Xie, Zheng, and Liu]{ding-etal-2021-nerd}
Ning Ding, Guangwei Xu, Yulin Chen, Xiaobin Wang, Xu~Han, Pengjun Xie, Haitao Zheng, and Zhiyuan Liu.
\newblock Few-{NERD}: A few-shot named entity recognition dataset.
\newblock In Chengqing Zong, Fei Xia, Wenjie Li, and Roberto Navigli (eds.), \emph{Proceedings of the 59th Annual Meeting of the Association for Computational Linguistics and the 11th International Joint Conference on Natural Language Processing (Volume 1: Long Papers)}, pp.\  3198--3213, Online, August 2021. Association for Computational Linguistics.
\newblock \doi{10.18653/v1/2021.acl-long.248}.
\newblock URL \url{https://aclanthology.org/2021.acl-long.248}.

\bibitem[Ding et~al.(2024)Ding, Zhang, Zhang, Xu, Shang, Xu, Yang, and Yang]{ding2024longrope}
Yiran Ding, Li~Lyna Zhang, Chengruidong Zhang, Yuanyuan Xu, Ning Shang, Jiahang Xu, Fan Yang, and Mao Yang.
\newblock Longrope: Extending llm context window beyond 2 million tokens.
\newblock \emph{arXiv preprint arXiv:2402.13753}, 2024.

\bibitem[Dong et~al.(2023)Dong, Li, Dai, Zheng, Wu, Chang, Sun, Xu, Li, and Sui]{dong2023survey}
Qingxiu Dong, Lei Li, Damai Dai, Ce~Zheng, Zhiyong Wu, Baobao Chang, Xu~Sun, Jingjing Xu, Lei Li, and Zhifang Sui.
\newblock A survey on in-context learning, 2023.

\bibitem[Fu et~al.(2024)Fu, Panda, Niu, Yue, Hajishirzi, Kim, and Peng]{fu2024data}
Yao Fu, Rameswar Panda, Xinyao Niu, Xiang Yue, Hannaneh Hajishirzi, Yoon Kim, and Hao Peng.
\newblock Data engineering for scaling language models to 128k context.
\newblock \emph{arXiv preprint arXiv:2402.10171}, 2024.

\bibitem[Gu \& Dao(2023)Gu and Dao]{gu2023mamba}
Albert Gu and Tri Dao.
\newblock Mamba: Linear-time sequence modeling with selective state spaces.
\newblock \emph{arXiv preprint arXiv:2312.00752}, 2023.

\bibitem[Hao et~al.(2022)Hao, Sun, Dong, Han, Gu, and Wei]{Hao2022StructuredPS}
Yaru Hao, Yutao Sun, Li~Dong, Zhixiong Han, Yuxian Gu, and Furu Wei.
\newblock Structured prompting: Scaling in-context learning to 1, 000 examples.
\newblock \emph{ArXiv}, abs/2212.06713, 2022.
\newblock URL \url{https://api.semanticscholar.org/CorpusID:254591686}.

\bibitem[Jiang et~al.(2023)Jiang, Sablayrolles, Mensch, Bamford, Chaplot, de~las Casas, Bressand, Lengyel, Lample, Saulnier, Lavaud, Lachaux, Stock, Scao, Lavril, Wang, Lacroix, and Sayed]{jiang2023mistral}
Albert~Q. Jiang, Alexandre Sablayrolles, Arthur Mensch, Chris Bamford, Devendra~Singh Chaplot, Diego de~las Casas, Florian Bressand, Gianna Lengyel, Guillaume Lample, Lucile Saulnier, Lélio~Renard Lavaud, Marie-Anne Lachaux, Pierre Stock, Teven~Le Scao, Thibaut Lavril, Thomas Wang, Timothée Lacroix, and William~El Sayed.
\newblock Mistral 7b, 2023.

\bibitem[Jin et~al.(2024)Jin, Han, Yang, Jiang, Liu, Chang, Chen, and Hu]{jin2024llm}
Hongye Jin, Xiaotian Han, Jingfeng Yang, Zhimeng Jiang, Zirui Liu, Chia-Yuan Chang, Huiyuan Chen, and Xia Hu.
\newblock Llm maybe longlm: Self-extend llm context window without tuning, 2024.

\bibitem[Kenton \& Toutanova(2019)Kenton and Toutanova]{kenton2019bert}
Jacob Devlin Ming-Wei~Chang Kenton and Lee~Kristina Toutanova.
\newblock Bert: Pre-training of deep bidirectional transformers for language understanding.
\newblock In \emph{Proceedings of NAACL-HLT}, pp.\  4171--4186, 2019.

\bibitem[Levy et~al.(2024)Levy, Jacoby, and Goldberg]{levy2024task}
Mosh Levy, Alon Jacoby, and Yoav Goldberg.
\newblock Same task, more tokens: the impact of input length on the reasoning performance of large language models, 2024.

\bibitem[Li et~al.(2023{\natexlab{a}})Li, Shao, Xie, Sheng, Zheng, Gonzalez, Stoica, Ma, and Zhang]{li2023how}
Dacheng Li, Rulin Shao, Anze Xie, Ying Sheng, Lianmin Zheng, Joseph Gonzalez, Ion Stoica, Xuezhe Ma, and Hao Zhang.
\newblock How long can context length of open-source {LLM}s truly promise?
\newblock In \emph{NeurIPS 2023 Workshop on Instruction Tuning and Instruction Following}, 2023{\natexlab{a}}.
\newblock URL \url{https://openreview.net/forum?id=LywifFNXV5}.

\bibitem[Li et~al.(2023{\natexlab{b}})Li, Wang, Zheng, and Zhang]{li2023loogle}
Jiaqi Li, Mengmeng Wang, Zilong Zheng, and Muhan Zhang.
\newblock Loogle: Can long-context language models understand long contexts?, 2023{\natexlab{b}}.

\bibitem[Li et~al.(2023{\natexlab{c}})Li, Gong, Feng, Xu, Zhang, Wu, and Kong]{li2023incontext}
Mukai Li, Shansan Gong, Jiangtao Feng, Yiheng Xu, Jun Zhang, Zhiyong Wu, and Lingpeng Kong.
\newblock In-context learning with many demonstration examples, 2023{\natexlab{c}}.

\bibitem[Liu et~al.(2022)Liu, Shen, Zhang, Dolan, Carin, and Chen]{liu-etal-2022-makes}
Jiachang Liu, Dinghan Shen, Yizhe Zhang, Bill Dolan, Lawrence Carin, and Weizhu Chen.
\newblock What makes good in-context examples for {GPT}-3?
\newblock In Eneko Agirre, Marianna Apidianaki, and Ivan Vuli{\'c} (eds.), \emph{Proceedings of Deep Learning Inside Out (DeeLIO 2022): The 3rd Workshop on Knowledge Extraction and Integration for Deep Learning Architectures}, pp.\  100--114, Dublin, Ireland and Online, May 2022. Association for Computational Linguistics.
\newblock \doi{10.18653/v1/2022.deelio-1.10}.
\newblock URL \url{https://aclanthology.org/2022.deelio-1.10}.

\bibitem[Liu et~al.(2024)Liu, Bai, Zhang, Zhang, Zhang, Zhang, Wang, Que, Chen, Su, et~al.]{liu20242}
Jiaheng Liu, Zhiqi Bai, Yuanxing Zhang, Chenchen Zhang, Yu~Zhang, Ge~Zhang, Jiakai Wang, Haoran Que, Yukang Chen, Wenbo Su, et~al.
\newblock E\^{} 2-llm: Efficient and extreme length extension of large language models.
\newblock \emph{arXiv preprint arXiv:2401.06951}, 2024.

\bibitem[Liu et~al.(2023)Liu, Lin, Hewitt, Paranjape, Bevilacqua, Petroni, and Liang]{Liu2023LostIT}
Nelson~F. Liu, Kevin Lin, John Hewitt, Ashwin Paranjape, Michele Bevilacqua, Fabio Petroni, and Percy Liang.
\newblock Lost in the middle: How language models use long contexts.
\newblock \emph{Transactions of the Association for Computational Linguistics}, 12:\penalty0 157--173, 2023.
\newblock URL \url{https://api.semanticscholar.org/CorpusID:259360665}.

\bibitem[Milios et~al.(2023)Milios, Reddy, and Bahdanau]{milios2023incontext}
Aristides Milios, Siva Reddy, and Dzmitry Bahdanau.
\newblock In-context learning for text classification with many labels, 2023.

\bibitem[Mohtashami \& Jaggi(2023)Mohtashami and Jaggi]{mohtashami2023landmark}
Amirkeivan Mohtashami and Martin Jaggi.
\newblock Landmark attention: Random-access infinite context length for transformers.
\newblock In \emph{Workshop on Efficient Systems for Foundation Models@ ICML2023}, 2023.

\bibitem[Nallapati et~al.(2017)Nallapati, Zhai, and Zhou]{nallapati2017summarunner}
Ramesh Nallapati, Feifei Zhai, and Bowen Zhou.
\newblock Summarunner: A recurrent neural network based sequence model for extractive summarization of documents.
\newblock In \emph{Proceedings of the AAAI conference on artificial intelligence}, volume~31, 2017.

\bibitem[Orvieto et~al.(2023)Orvieto, Smith, Gu, Fernando, Gulcehre, Pascanu, and De]{Orvieto2023ResurrectingRN}
Antonio Orvieto, Samuel~L. Smith, Albert Gu, Anushan Fernando, Caglar Gulcehre, Razvan Pascanu, and Soham De.
\newblock Resurrecting recurrent neural networks for long sequences.
\newblock \emph{ArXiv}, abs/2303.06349, 2023.
\newblock URL \url{https://api.semanticscholar.org/CorpusID:257496654}.

\bibitem[Peng et~al.(2023{\natexlab{a}})Peng, Alcaide, Anthony, Albalak, Arcadinho, Biderman, Cao, Cheng, Chung, Derczynski, et~al.]{peng2023rwkv}
Bo~Peng, Eric Alcaide, Quentin Anthony, Alon Albalak, Samuel Arcadinho, Stella Biderman, Huanqi Cao, Xin Cheng, Michael Chung, Leon Derczynski, et~al.
\newblock Rwkv: Reinventing rnns for the transformer era.
\newblock In \emph{Findings of the Association for Computational Linguistics: EMNLP 2023}, pp.\  14048--14077, 2023{\natexlab{a}}.

\bibitem[Peng et~al.(2023{\natexlab{b}})Peng, Quesnelle, Fan, and Shippole]{peng2023yarn}
Bowen Peng, Jeffrey Quesnelle, Honglu Fan, and Enrico Shippole.
\newblock Yarn: Efficient context window extension of large language models, 2023{\natexlab{b}}.

\bibitem[Peng et~al.(2023{\natexlab{c}})Peng, Wang, Chen, Li, Qi, Wang, Wu, Zeng, Xu, Hou, and Li]{peng2023does}
Hao Peng, Xiaozhi Wang, Jianhui Chen, Weikai Li, Yunjia Qi, Zimu Wang, Zhili Wu, Kaisheng Zeng, Bin Xu, Lei Hou, and Juanzi Li.
\newblock When does in-context learning fall short and why? a study on specification-heavy tasks, 2023{\natexlab{c}}.

\bibitem[Press et~al.(2022)Press, Smith, and Lewis]{alibi}
Ofir Press, Noah Smith, and Mike Lewis.
\newblock Train short, test long: Attention with linear biases enables input length extrapolation.
\newblock In \emph{International Conference on Learning Representations}, 2022.
\newblock URL \url{https://openreview.net/forum?id=R8sQPpGCv0}.

\bibitem[Ratner et~al.(2023)Ratner, Levine, Belinkov, Ram, Magar, Abend, Karpas, Shashua, Leyton-Brown, and Shoham]{ratner-etal-2023-parallel}
Nir Ratner, Yoav Levine, Yonatan Belinkov, Ori Ram, Inbal Magar, Omri Abend, Ehud Karpas, Amnon Shashua, Kevin Leyton-Brown, and Yoav Shoham.
\newblock Parallel context windows for large language models.
\newblock In Anna Rogers, Jordan Boyd-Graber, and Naoaki Okazaki (eds.), \emph{Proceedings of the 61st Annual Meeting of the Association for Computational Linguistics (Volume 1: Long Papers)}, pp.\  6383--6402, Toronto, Canada, July 2023. Association for Computational Linguistics.
\newblock \doi{10.18653/v1/2023.acl-long.352}.
\newblock URL \url{https://aclanthology.org/2023.acl-long.352}.

\bibitem[Reid et~al.(2024)Reid, Savinov, Teplyashin, Lepikhin, Lillicrap, Alayrac, Soricut, Lazaridou, Firat, Schrittwieser, Antonoglou, Anil, Borgeaud, Dai, Millican, Dyer, Glaese, Sottiaux, Lee, Viola, Reynolds, Xu, Molloy, Chen, Isard, Barham, Hennigan, McIlroy, Johnson, Schalkwyk, Collins, Rutherford, Moreira, Ayoub, Goel, Meyer, Thornton, Yang, Michalewski, Abbas, Schucher, Anand, Ives, Keeling, Lenc, Haykal, Shakeri, Shyam, Chowdhery, Ring, Spencer, Sezener, Vilnis, Chang, Morioka, Tucker, Zheng, Woodman, Attaluri, Kocisky, Eltyshev, Chen, Chung, Selo, Brahma, Georgiev, Slone, Zhu, Lottes, Qiao, Caine, Riedel, Tomala, Chadwick, Love, Choy, Mittal, Houlsby, Tang, Lamm, Bai, Zhang, He, Cheng, Humphreys, Li, Brin, Cassirer, Miao, Zilka, Tobin, Xu, Proleev, Sohn, Magni, Hendricks, Gao, Ontan'on, Bunyan, Byrd, Sharma, Zhang, Pinto, Sinha, Mehta, Jia, Caelles, Webson, Morris, Roelofs, Ding, Strudel, Xiong, Ritter, Dehghani, Chaabouni, Karmarkar, Lai, Mentzer, Xu, Li, Zhang, Paine, Goldin, Neyshabur, Baumli,
  Levskaya, Laskin, Jia, Rae, Xiao, He, Giordano, Yagati, Lespiau, Natsev, Ganapathy, Liu, Martins, Chen, Xu, Barnes, May, Vezer, Oh, Franko, Bridgers, Zhao, Wu, Mustafa, Sechrist, Parisotto, Pillai, Larkin, Gu, Sorokin, Krikun, Guseynov, Landon, Datta, Pritzel, Thacker, Yang, Hui, Hauth, Yeh, Barker, Mao-Jones, Austin, Sheahan, Schuh, Svensson, Jain, Ramasesh, Briukhov, Chung, von Glehn, Butterfield, Jhakra, Wiethoff, Frye, Grimstad, Changpinyo, Lan, Bortsova, Wu, Voigtlaender, Sainath, Smith, Hawkins, Cao, Besley, Srinivasan, Omernick, Gaffney, de~Castro~Surita, Burnell, Damoc, Ahn, Brock, Pajarskas, Petrushkina, Noury, Blanco, Swersky, Ahuja, Avrahami, Misra, de~Liedekerke, Iinuma, Polozov, York, van~den Driessche, Michel, Chiu, Blevins, Gleicher, Recasens, Rrustemi, Gribovskaya, Roy, Gworek, Arnold, Lee, Lee-Thorp, Maggioni, Piqueras, Badola, Vikram, Gonzalez, Baddepudi, Senter, Devlin, Qin, Azzam, Trebacz, Polacek, Krishnakumar, yiin Chang, Tung, Penchev, Joshi, Olszewska, Muir, Wirth, Hartman, Newlan,
  Kashem, Bolina, Dabir, van Amersfoort, Ahmed, Cobon-Kerr, Kamath, Hrafnkelsson, Hou, Mackinnon, Frechette, Noland, Si, Taropa, Li, Crone, Gulati, Cevey, Adler, Ma, Silver, Tokumine, Powell, Lee, Chang, Hassan, Mincu, Yang, Levine, Brennan, Wang, Hodkinson, Zhao, Lipschultz, Pope, Chang, Li, Shafey, Paganini, Douglas, Bohnet, Pardo, Odoom, Rosca, dos Santos, Soparkar, Guez, Hudson, Hansen, Asawaroengchai, Addanki, Yu, Stokowiec, Khan, Gilmer, Lee, Bostock, Rong, Caton, Pejman, Pavetic, Brown, Sharma, Luvci'c, Samuel, Djolonga, Mandhane, Sjosund, Buchatskaya, White, Clay, Jiang, Lim, Hemsley, Labanowski, Cao, Steiner, Hashemi, Austin, Gergely, Blyth, Stanton, Shivakumar, Siddhant, Andreassen, Araya, Sethi, Shivanna, Hand, Bapna, Khodaei, Miech, Tanzer, Swing, Thakoor, Pan, Nado, Winkler, Yu, Saleh, Maggiore, Barr, Giang, Kagohara, Danihelka, Marathe, Feinberg, Elhawaty, Ghelani, Horgan, Miller, Walker, Tanburn, Tariq, Shrivastava, Xia, Chiu, Ashwood, Baatarsukh, Samangooei, Alcober, Stjerngren, Komarek,
  Tsihlas, Boral, Comanescu, Chen, Liu, Bloxwich, Chen, Sun, Feng, Mauger, Dotiwalla, Hellendoorn, Sharman, Zheng, Haridasan, Barth-Maron, Swanson, Rogozi'nska, Andreev, Rubenstein, Sang, Hurt, Elsayed, Wang, Lacey, Ili'c, Zhao, Aroyo, Iwuanyanwu, Nikolaev, Lakshminarayanan, Jazayeri, Kaufman, Varadarajan, Tekur, Fritz, Khalman, Reitter, Dasgupta, Sarcar, Ornduff, Snaider, Huot, Jia, Kemp, Trdin, Vijayakumar, Kim, Angermueller, Lao, Liu, Zhang, Engel, Greene, White, Austin, Taylor, Ashraf, Liu, Georgaki, Cai, Kulizhskaya, Goenka, Saeta, Vodrahalli, Frank, de~Cesare, Robenek, Richardson, Alnahlawi, Yew, Ponnapalli, Tagliasacchi, Korchemniy, Kim, Li, Rosgen, Levin, Wiesner, Banzal, Srinivasan, Yu, cCauglar Unlu, Reid, Tung, Finchelstein, Kumar, Elisseeff, Huang, Zhang, Zhu, Aguilar, Gim'enez, Xia, Dousse, Gierke, Yeganeh, Yates, Jalan, Li, Latorre-Chimoto, Nguyen, Durden, Kallakuri, Liu, Johnson, Tsai, Talbert, Liu, Neitz, Elkind, Selvi, Jasarevic, Soares, Cui, Wang, Wang, Ye, Kallarackal, Loher, Lam, Broder,
  Holtmann-Rice, Martin, Ramadhana, Toyama, Shukla, Basu, Mohan, Fernando, Fiedel, Paterson, Li, Garg, Park, Choi, Wu, Singh, Zhang, Globerson, Yu, Carpenter, de~Chaumont~Quitry, Radebaugh, Lin, Tudor, Shroff, Garmon, Du, Vats, Lu, Iqbal, Yakubovich, Tripuraneni, Manyika, Qureshi, Hua, Ngani, Raad, Forbes, Bulanova, Stanway, Sundararajan, Ungureanu, Bishop, Li, Venkatraman, Li, Thornton, Scellato, Gupta, Wang, Tenney, Wu, Shenoy, Carvajal, Wright, Bariach, Xiao, Hawkins, Dalmia, Farabet, Valenzuela, Yuan, Welty, Agarwal, Chen, Kim, Hulse, Dukkipati, Paszke, Bolt, Davoodi, Choo, Beattie, Prendki, Vashisht, Santamaria-Fernandez, Cobo, Wilkiewicz, Madras, Elqursh, Uy, Ramirez, Harvey, Liechty, Zen, Seibert, Hu, Khorlin, Le, Aharoni, Li, Wang, Kumar, Lince, Casagrande, Hoover, Badawy, Soergel, Vnukov, Miecnikowski, Simsa, Koop, Kumar, Sellam, Vlasic, Daruki, Shabat, Zhang, Su, Zhang, Liu, Sun, Palmer, Ghaffarkhah, Xiong, Cotruta, Fink, Dixon, Sreevatsa, Goedeckemeyer, Dimitriev, Jafari, Crocker, Fitzgerald,
  Kumar, Ghemawat, Philips, Liu, Liang, Sterneck, Repina, Wu, Knight, Georgiev, Lee, Askham, Chakladar, Louis, Crous, Cate, Petrova, Quinn, Owusu-Afriyie, Singhal, Wei, Kim, Vincent, Nasr, Choquette-Choo, Tojo, Lu, de~Las~Casas, Cheng, Bolukbasi, Lee, Fatehi, Ananthanarayanan, Patel, Kaed, Li, Sygnowski, Belle, Chen, Konzelmann, Poder, Garg, Koverkathu, Brown, Dyer, Liu, Nova, Xu, Petrov, Hassabis, Kavukcuoglu, Dean, and Vinyals]{Reid2024Gemini1U}
Machel Reid, Nikolay Savinov, Denis Teplyashin, Dmitry Lepikhin, Timothy~P. Lillicrap, Jean-Baptiste Alayrac, Radu Soricut, Angeliki Lazaridou, Orhan Firat, Julian Schrittwieser, Ioannis Antonoglou, Rohan Anil, Sebastian Borgeaud, Andrew~M. Dai, Katie Millican, Ethan Dyer, Mia Glaese, Thibault Sottiaux, Benjamin Lee, Fabio Viola, Malcolm Reynolds, Yuanzhong Xu, James Molloy, Jilin Chen, Michael Isard, Paul Barham, Tom Hennigan, Ross McIlroy, Melvin Johnson, Johan Schalkwyk, Eli Collins, Eliza Rutherford, Erica Moreira, Kareem~W. Ayoub, Megha Goel, Clemens Meyer, Gregory Thornton, Zhen Yang, Henryk Michalewski, Zaheer Abbas, Nathan Schucher, Ankesh Anand, Richard Ives, James Keeling, Karel Lenc, Salem Haykal, Siamak Shakeri, Pranav Shyam, Aakanksha Chowdhery, Roman Ring, Stephen Spencer, Eren Sezener, Luke Vilnis, Oscar Chang, Nobuyuki Morioka, George Tucker, Ce~Zheng, Oliver Woodman, Nithya Attaluri, Tomas Kocisky, Evgenii Eltyshev, Xi~Chen, Timothy Chung, Vittorio Selo, Siddhartha Brahma, Petko Georgiev,
  Ambrose Slone, Zhenkai Zhu, James Lottes, Siyuan Qiao, Ben Caine, Sebastian Riedel, Alex Tomala, Martin Chadwick, J~Christopher Love, Peter Choy, Sid Mittal, Neil Houlsby, Yunhao Tang, Matthew Lamm, Libin Bai, Qiao Zhang, Luheng He, Yong Cheng, Peter Humphreys, Yujia Li, Sergey Brin, Albin Cassirer, Ying-Qi Miao, Lukas Zilka, Taylor Tobin, Kelvin Xu, Lev Proleev, Daniel Sohn, Alberto Magni, Lisa~Anne Hendricks, Isabel Gao, Santiago Ontan'on, Oskar Bunyan, Nathan Byrd, Abhanshu Sharma, Biao Zhang, Mario Pinto, Rishika Sinha, Harsh Mehta, Dawei Jia, Sergi Caelles, Albert Webson, Alex Morris, Becca Roelofs, Yifan Ding, Robin Strudel, Xuehan Xiong, Marvin Ritter, Mostafa Dehghani, Rahma Chaabouni, Abhijit Karmarkar, Guangda Lai, Fabian Mentzer, Bibo Xu, YaGuang Li, Yujing Zhang, Tom~Le Paine, Alex Goldin, Behnam Neyshabur, Kate Baumli, Anselm Levskaya, Michael Laskin, Wenhao Jia, Jack~W. Rae, Kefan Xiao, Antoine He, Skye Giordano, Lakshman Yagati, Jean-Baptiste Lespiau, Paul Natsev, Sanjay Ganapathy, Fangyu
  Liu, Danilo Martins, Nanxin Chen, Yunhan Xu, Megan Barnes, Rhys May, Arpi Vezer, Junhyuk Oh, Ken Franko, Sophie Bridgers, Ruizhe Zhao, Boxi Wu, Basil Mustafa, Sean Sechrist, Emilio Parisotto, Thanumalayan~Sankaranarayana Pillai, Chris Larkin, Chenjie Gu, Christina Sorokin, Maxim Krikun, Alexey Guseynov, Jessica Landon, Romina Datta, Alexander Pritzel, Phoebe Thacker, Fan Yang, Kevin Hui, A.E. Hauth, Chih-Kuan Yeh, David Barker, Justin Mao-Jones, Sophia Austin, Hannah Sheahan, Parker Schuh, James Svensson, Rohan Jain, Vinay~Venkatesh Ramasesh, Anton Briukhov, Da-Woon Chung, Tamara von Glehn, Christina Butterfield, Priya Jhakra, Matt Wiethoff, Justin Frye, Jordan Grimstad, Beer Changpinyo, Charline~Le Lan, Anna Bortsova, Yonghui Wu, Paul Voigtlaender, Tara~N. Sainath, Charlotte Smith, Will Hawkins, Kris Cao, James Besley, Srivatsan Srinivasan, Mark Omernick, Colin Gaffney, Gabriela de~Castro~Surita, Ryan Burnell, Bogdan Damoc, Junwhan Ahn, Andrew Brock, Mantas Pajarskas, Anastasia Petrushkina, Seb Noury,
  Lorenzo Blanco, Kevin Swersky, Arun Ahuja, Thi Avrahami, Vedant Misra, Raoul de~Liedekerke, Mariko Iinuma, Alex Polozov, Sarah York, George van~den Driessche, Paul Michel, Justin Chiu, Rory Blevins, Zach Gleicher, Adria Recasens, Alban Rrustemi, Elena Gribovskaya, Aurko Roy, Wiktor Gworek, S'ebastien M.~R. Arnold, Lisa Lee, James Lee-Thorp, Marcello Maggioni, Enrique Piqueras, Kartikeya Badola, Sharad Vikram, Lucas Gonzalez, Anirudh Baddepudi, Evan Senter, Jacob Devlin, James Qin, Michael Azzam, Maja Trebacz, Martin Polacek, Kashyap Krishnakumar, Shuo yiin Chang, Matthew Tung, Ivo Penchev, Rishabh Joshi, Kate Olszewska, Carrie Muir, Mateo Wirth, Ale~Jakse Hartman, Joshua Newlan, Sheleem Kashem, Vijay Bolina, Elahe Dabir, Joost~R. van Amersfoort, Zafarali Ahmed, James Cobon-Kerr, Aishwarya~B Kamath, Arnar~Mar Hrafnkelsson, Le~Hou, Ian Mackinnon, Alexandre Frechette, Eric Noland, Xiance Si, Emanuel Taropa, Dong Li, Phil Crone, Anmol Gulati, S'ebastien Cevey, Jonas Adler, Ada Ma, David Silver, Simon Tokumine,
  Richard Powell, Stephan Lee, Michael~B. Chang, Samer Hassan, Diana Mincu, Antoine Yang, Nir Levine, Jenny Brennan, Mingqiu Wang, Sarah Hodkinson, Jeffrey Zhao, Josh Lipschultz, Aedan Pope, Michael~B. Chang, Cheng Li, Laurent~El Shafey, Michela Paganini, Sholto Douglas, Bernd Bohnet, Fabio Pardo, Seth Odoom, Mihaela Rosca, Cicero~Nogueira dos Santos, Kedar Soparkar, Arthur Guez, Tom Hudson, Steven Hansen, Chulayuth Asawaroengchai, Ravichandra Addanki, Tianhe Yu, Wojciech Stokowiec, Mina Khan, Justin Gilmer, Jaehoon Lee, Carrie~Grimes Bostock, Keran Rong, Jonathan Caton, Pedram Pejman, Filip Pavetic, Geoff Brown, Vivek Sharma, Mario Luvci'c, Rajkumar Samuel, Josip Djolonga, Amol Mandhane, Lars~Lowe Sjosund, Elena Buchatskaya, Elspeth White, Natalie Clay, Jiepu Jiang, Hyeontaek Lim, Ross Hemsley, Jane Labanowski, Nicola~De Cao, David Steiner, Sayed~Hadi Hashemi, Jacob Austin, Anita Gergely, Tim Blyth, Joe Stanton, Kaushik Shivakumar, Aditya Siddhant, Anders Andreassen, Carlos~L. Araya, Nikhil Sethi, Rakesh
  Shivanna, Steven Hand, Ankur Bapna, Ali Khodaei, Antoine Miech, Garrett Tanzer, Andy Swing, Shantanu Thakoor, Zhufeng Pan, Zachary Nado, Stephanie Winkler, Dian Yu, Mohammad Saleh, Lorenzo Maggiore, Iain Barr, Minh Giang, Thais Kagohara, Ivo Danihelka, Amit Marathe, Vladimir Feinberg, Mohamed Elhawaty, Nimesh Ghelani, Dan Horgan, Helen Miller, Lexi Walker, Richard Tanburn, Mukarram Tariq, Disha Shrivastava, Fei Xia, Chung-Cheng Chiu, Zoe~C. Ashwood, Khuslen Baatarsukh, Sina Samangooei, Fred Alcober, Axel Stjerngren, Paul Komarek, Katerina Tsihlas, Anudhyan Boral, Ramona Comanescu, Jeremy Chen, Ruibo Liu, Dawn Bloxwich, Charlie Chen, Yanhua Sun, Fangxiaoyu Feng, Matthew Mauger, Xerxes Dotiwalla, Vincent Hellendoorn, Michael Sharman, Ivy Zheng, Krishna Haridasan, Gabriel Barth-Maron, Craig Swanson, Dominika Rogozi'nska, Alek Andreev, Paul~Kishan Rubenstein, Ruoxin Sang, Dan Hurt, Gamaleldin Elsayed, Renshen Wang, Dave Lacey, Anastasija Ili'c, Yao Zhao, Lora Aroyo, Chimezie Iwuanyanwu, Vitaly Nikolaev, Balaji
  Lakshminarayanan, Sadegh Jazayeri, Raphael~Lopez Kaufman, Mani Varadarajan, Chetan Tekur, Doug Fritz, Misha Khalman, David Reitter, Kingshuk Dasgupta, Shourya Sarcar, T.~Ornduff, Javier Snaider, Fantine Huot, Johnson Jia, Rupert Kemp, Nejc Trdin, Anitha Vijayakumar, Lucy Kim, Christof Angermueller, Li~Lao, Tianqi Liu, Haibin Zhang, David Engel, Somer Greene, Anais White, Jessica Austin, Lilly Taylor, Shereen Ashraf, Dangyi Liu, Maria Georgaki, Irene Cai, Yana Kulizhskaya, Sonam Goenka, Brennan Saeta, Kiran Vodrahalli, Christian Frank, Dario de~Cesare, Brona Robenek, Harry Richardson, Mahmoud Alnahlawi, Christopher Yew, Priya Ponnapalli, Marco Tagliasacchi, Alex Korchemniy, Yelin Kim, Dinghua Li, Bill Rosgen, Kyle Levin, Jeremy Wiesner, Praseem Banzal, Praveen Srinivasan, Hongkun Yu, cCauglar Unlu, David Reid, Zora Tung, Daniel~F. Finchelstein, Ravin Kumar, Andre Elisseeff, Jin Huang, Ming Zhang, Rui Zhu, Ricardo Aguilar, Mai Gim'enez, Jiawei Xia, Olivier Dousse, Willi Gierke, Soheil~Hassas Yeganeh, Damion
  Yates, Komal Jalan, Lu~Li, Eri Latorre-Chimoto, Duc~Dung Nguyen, Ken Durden, Praveen Kallakuri, Yaxin Liu, Matthew Johnson, Tomy Tsai, Alice Talbert, Jasmine Liu, Alexander Neitz, Chen Elkind, Marco Selvi, Mimi Jasarevic, Livio~Baldini Soares, Albert Cui, Pidong Wang, Alek~Wenjiao Wang, Xinyu Ye, Krystal Kallarackal, Lucia Loher, Hoi Lam, Josef Broder, Daniel~Niels Holtmann-Rice, Nina Martin, Bramandia Ramadhana, Daniel Toyama, Mrinal Shukla, Sujoy Basu, Abhi Mohan, Nicholas Fernando, Noah Fiedel, Kim Paterson, Hui Li, Ankush Garg, Jane Park, Donghyun Choi, Diane Wu, Sankalp Singh, Zhishuai Zhang, Amir Globerson, Lily Yu, John Carpenter, F{\'e}lix de~Chaumont~Quitry, Carey Radebaugh, Chu-Cheng Lin, Alex Tudor, Prakash Shroff, Drew Garmon, Dayou Du, Neera Vats, Han Lu, Shariq Iqbal, Alexey Yakubovich, Nilesh Tripuraneni, James Manyika, Haroon Qureshi, Nan Hua, Christel Ngani, Maria~Abi Raad, Hannah Forbes, Anna Bulanova, Jeff Stanway, Mukund Sundararajan, Victor Ungureanu, Colton Bishop, Yunjie Li, Balaji
  Venkatraman, Bo~Li, Chloe Thornton, Salvatore Scellato, Nishesh Gupta, Yicheng Wang, Ian Tenney, Xihui Wu, Ashish Shenoy, Gabriel Carvajal, Diana~Gage Wright, Ben Bariach, Zhuyun Xiao, Peter Hawkins, Sid Dalmia, Cl'ement Farabet, Pedro Valenzuela, Quan Yuan, Christoper~A. Welty, Ananth Agarwal, Mianna Chen, Wooyeol Kim, Brice Hulse, Nandita Dukkipati, Adam Paszke, Andrew Bolt, Elnaz Davoodi, Kiam Choo, Jennifer Beattie, Jennifer Prendki, Harsha Vashisht, Rebeca Santamaria-Fernandez, Luis~C. Cobo, Jarek Wilkiewicz, David Madras, Ali Elqursh, Grant Uy, Kevin Ramirez, Matt Harvey, Tyler Liechty, Heiga Zen, Jeff Seibert, Clara~Huiyi Hu, A.~Ya. Khorlin, Maigo Le, Asaf Aharoni, Megan Li, Lily Wang, Sandeep Kumar, Alejandro Lince, Norman Casagrande, Jay Hoover, Dalia~El Badawy, David Soergel, Denis Vnukov, Matt Miecnikowski, Jiři Simsa, Anna Koop, Praveen Kumar, Thibault Sellam, Daniel Vlasic, Samira Daruki, Nir Shabat, John Zhang, Guolong Su, Jiageng Zhang, Jeremiah Liu, Yi~Sun, Evan Palmer, Alireza Ghaffarkhah,
  Xi~Xiong, Victor Cotruta, Michael Fink, Lucas Dixon, Ashwin Sreevatsa, Adrian Goedeckemeyer, Alek Dimitriev, Mohsen Jafari, Remi Crocker, Nicholas~A Fitzgerald, Aviral Kumar, Sanjay Ghemawat, Ivan Philips, Frederick Liu, Yannie Liang, Rachel Sterneck, Alena Repina, Marcus Wu, Laura Knight, Marin Georgiev, Hyo Lee, Harry Askham, Abhishek Chakladar, Annie Louis, Carl Crous, Hardie Cate, Dessie Petrova, Michael Quinn, Denese Owusu-Afriyie, Achintya Singhal, Nan Wei, Solomon Kim, Damien Vincent, Milad Nasr, Christopher~A. Choquette-Choo, Reiko Tojo, Shawn Lu, Diego de~Las~Casas, Yuchung Cheng, Tolga Bolukbasi, Katherine Lee, Saaber Fatehi, Rajagopal Ananthanarayanan, Miteyan Patel, Charbel~El Kaed, Jing Li, Jakub Sygnowski, Shreyas~Rammohan Belle, Zhe Chen, Jaclyn Konzelmann, Siim Poder, Roopal Garg, Vinod Koverkathu, Adam Brown, Chris Dyer, Rosanne Liu, Azade Nova, Jun Xu, Slav Petrov, Demis Hassabis, Koray Kavukcuoglu, Jeffrey Dean, and Oriol Vinyals.
\newblock Gemini 1.5: Unlocking multimodal understanding across millions of tokens of context.
\newblock \emph{ArXiv}, abs/2403.05530, 2024.
\newblock URL \url{https://api.semanticscholar.org/CorpusID:268297180}.

\bibitem[Rozière et~al.(2024)Rozière, Gehring, Gloeckle, Sootla, Gat, Tan, Adi, Liu, Sauvestre, Remez, Rapin, Kozhevnikov, Evtimov, Bitton, Bhatt, Ferrer, Grattafiori, Xiong, Défossez, Copet, Azhar, Touvron, Martin, Usunier, Scialom, and Synnaeve]{rozière2024code}
Baptiste Rozière, Jonas Gehring, Fabian Gloeckle, Sten Sootla, Itai Gat, Xiaoqing~Ellen Tan, Yossi Adi, Jingyu Liu, Romain Sauvestre, Tal Remez, Jérémy Rapin, Artyom Kozhevnikov, Ivan Evtimov, Joanna Bitton, Manish Bhatt, Cristian~Canton Ferrer, Aaron Grattafiori, Wenhan Xiong, Alexandre Défossez, Jade Copet, Faisal Azhar, Hugo Touvron, Louis Martin, Nicolas Usunier, Thomas Scialom, and Gabriel Synnaeve.
\newblock Code llama: Open foundation models for code, 2024.

\bibitem[Sileo et~al.(2019)Sileo, Van De~Cruys, Pradel, and Muller]{sileo-etal-2019-mining}
Damien Sileo, Tim Van De~Cruys, Camille Pradel, and Philippe Muller.
\newblock Mining discourse markers for unsupervised sentence representation learning.
\newblock In Jill Burstein, Christy Doran, and Thamar Solorio (eds.), \emph{Proceedings of the 2019 Conference of the North {A}merican Chapter of the Association for Computational Linguistics: Human Language Technologies, Volume 1 (Long and Short Papers)}, pp.\  3477--3486, Minneapolis, Minnesota, June 2019. Association for Computational Linguistics.
\newblock \doi{10.18653/v1/N19-1351}.
\newblock URL \url{https://aclanthology.org/N19-1351}.

\bibitem[Su et~al.(2021)Su, Lu, Pan, Wen, and Liu]{Su2021RoFormerET}
Jianlin Su, Yu~Lu, Shengfeng Pan, Bo~Wen, and Yunfeng Liu.
\newblock Roformer: Enhanced transformer with rotary position embedding.
\newblock \emph{ArXiv}, abs/2104.09864, 2021.
\newblock URL \url{https://api.semanticscholar.org/CorpusID:233307138}.

\bibitem[Su et~al.(2024)Su, Ahmed, Lu, Pan, Bo, and Liu]{su2024roformer}
Jianlin Su, Murtadha Ahmed, Yu~Lu, Shengfeng Pan, Wen Bo, and Yunfeng Liu.
\newblock Roformer: Enhanced transformer with rotary position embedding.
\newblock \emph{Neurocomputing}, 568:\penalty0 127063, 2024.

\bibitem[Tay et~al.(2021)Tay, Dehghani, Abnar, Shen, Bahri, Pham, Rao, Yang, Ruder, and Metzler]{tay2021long}
Yi~Tay, Mostafa Dehghani, Samira Abnar, Yikang Shen, Dara Bahri, Philip Pham, Jinfeng Rao, Liu Yang, Sebastian Ruder, and Donald Metzler.
\newblock Long range arena : A benchmark for efficient transformers.
\newblock In \emph{International Conference on Learning Representations}, 2021.
\newblock URL \url{https://openreview.net/forum?id=qVyeW-grC2k}.

\bibitem[Team et~al.(2023)Team, Anil, Borgeaud, Wu, Alayrac, Yu, Soricut, Schalkwyk, Dai, Hauth, et~al.]{team2023gemini}
Gemini Team, Rohan Anil, Sebastian Borgeaud, Yonghui Wu, Jean-Baptiste Alayrac, Jiahui Yu, Radu Soricut, Johan Schalkwyk, Andrew~M Dai, Anja Hauth, et~al.
\newblock Gemini: a family of highly capable multimodal models.
\newblock \emph{arXiv preprint arXiv:2312.11805}, 2023.

\bibitem[Team et~al.(2024)Team, Mesnard, Hardin, Dadashi, Bhupatiraju, Pathak, Sifre, Rivi{\`e}re, Kale, Love, et~al.]{team2024gemma}
Gemma Team, Thomas Mesnard, Cassidy Hardin, Robert Dadashi, Surya Bhupatiraju, Shreya Pathak, Laurent Sifre, Morgane Rivi{\`e}re, Mihir~Sanjay Kale, Juliette Love, et~al.
\newblock Gemma: Open models based on gemini research and technology.
\newblock \emph{arXiv preprint arXiv:2403.08295}, 2024.

\bibitem[Tworkowski et~al.(2023)Tworkowski, Staniszewski, Pacek, Wu, Michalewski, and Miłoś]{tworkowski2023focused}
Szymon Tworkowski, Konrad Staniszewski, Mikołaj Pacek, Yuhuai Wu, Henryk Michalewski, and Piotr Miłoś.
\newblock Focused transformer: Contrastive training for context scaling, 2023.

\bibitem[Vuli{\'c} et~al.(2021)Vuli{\'c}, Su, Coope, Gerz, Budzianowski, Casanueva, Mrk{\v{s}}i{\'c}, and Wen]{vulic-etal-2021-convfit}
Ivan Vuli{\'c}, Pei-Hao Su, Samuel Coope, Daniela Gerz, Pawe{\l} Budzianowski, I{\~n}igo Casanueva, Nikola Mrk{\v{s}}i{\'c}, and Tsung-Hsien Wen.
\newblock {ConvFiT:} {C}onversational fine-tuning of pretrained language models.
\newblock In Marie-Francine Moens, Xuanjing Huang, Lucia Specia, and Scott Wen-tau Yih (eds.), \emph{Proceedings of the 2021 Conference on Empirical Methods in Natural Language Processing}, pp.\  1151--1168, Online and Punta Cana, Dominican Republic, November 2021. Association for Computational Linguistics.
\newblock \doi{10.18653/v1/2021.emnlp-main.88}.
\newblock URL \url{https://aclanthology.org/2021.emnlp-main.88}.

\bibitem[Wang et~al.(2023)Wang, Dong, Cheng, Liu, Yan, Gao, and Wei]{wang2023augmenting}
Weizhi Wang, Li~Dong, Hao Cheng, Xiaodong Liu, Xifeng Yan, Jianfeng Gao, and Furu Wei.
\newblock Augmenting language models with long-term memory.
\newblock In \emph{Thirty-seventh Conference on Neural Information Processing Systems}, 2023.
\newblock URL \url{https://openreview.net/forum?id=BryMFPQ4L6}.

\bibitem[Wu et~al.(2023)Wu, Wang, Ye, and Kong]{wu2023selfadaptive}
Zhiyong Wu, Yaoxiang Wang, Jiacheng Ye, and Lingpeng Kong.
\newblock Self-adaptive in-context learning: An information compression perspective for in-context example selection and ordering, 2023.

\bibitem[Xiao et~al.(2024)Xiao, Tian, Chen, Han, and Lewis]{xiao2024efficient}
Guangxuan Xiao, Yuandong Tian, Beidi Chen, Song Han, and Mike Lewis.
\newblock Efficient streaming language models with attention sinks.
\newblock In \emph{The Twelfth International Conference on Learning Representations}, 2024.
\newblock URL \url{https://openreview.net/forum?id=NG7sS51zVF}.

\bibitem[Xiong et~al.(2023)Xiong, Liu, Molybog, Zhang, Bhargava, Hou, Martin, Rungta, Sankararaman, Oguz, et~al.]{xiong2023effective}
Wenhan Xiong, Jingyu Liu, Igor Molybog, Hejia Zhang, Prajjwal Bhargava, Rui Hou, Louis Martin, Rashi Rungta, Karthik~Abinav Sankararaman, Barlas Oguz, et~al.
\newblock Effective long-context scaling of foundation models.
\newblock \emph{arXiv preprint arXiv:2309.16039}, 2023.

\bibitem[Yu et~al.(2020)Yu, Sun, Cardie, and Yu]{yu-etal-2020-dialogue}
Dian Yu, Kai Sun, Claire Cardie, and Dong Yu.
\newblock Dialogue-based relation extraction.
\newblock In Dan Jurafsky, Joyce Chai, Natalie Schluter, and Joel Tetreault (eds.), \emph{Proceedings of the 58th Annual Meeting of the Association for Computational Linguistics}, pp.\  4927--4940, Online, July 2020. Association for Computational Linguistics.
\newblock \doi{10.18653/v1/2020.acl-main.444}.
\newblock URL \url{https://aclanthology.org/2020.acl-main.444}.

\bibitem[Zeng et~al.(2022)Zeng, Liu, Du, Wang, Lai, Ding, Yang, Xu, Zheng, Xia, et~al.]{zeng2022glm}
Aohan Zeng, Xiao Liu, Zhengxiao Du, Zihan Wang, Hanyu Lai, Ming Ding, Zhuoyi Yang, Yifan Xu, Wendi Zheng, Xiao Xia, et~al.
\newblock Glm-130b: An open bilingual pre-trained model.
\newblock In \emph{The Eleventh International Conference on Learning Representations}, 2022.

\bibitem[Zhang et~al.(2024)Zhang, Chen, Hu, Xu, Chen, Hao, Han, Thai, Wang, Liu, and Sun]{zhang2024inftybench}
Xinrong Zhang, Yingfa Chen, Shengding Hu, Zihang Xu, Junhao Chen, Moo~Khai Hao, Xu~Han, Zhen~Leng Thai, Shuo Wang, Zhiyuan Liu, and Maosong Sun.
\newblock $\infty$bench: Extending long context evaluation beyond 100k tokens, 2024.

\bibitem[Zhang et~al.(2017)Zhang, Zhong, Chen, Angeli, and Manning]{zhang2017tacred}
Yuhao Zhang, Victor Zhong, Danqi Chen, Gabor Angeli, and Christopher~D. Manning.
\newblock Position-aware attention and supervised data improve slot filling.
\newblock In \emph{Proceedings of the 2017 Conference on Empirical Methods in Natural Language Processing (EMNLP 2017)}, pp.\  35--45, 2017.
\newblock URL \url{https://nlp.stanford.edu/pubs/zhang2017tacred.pdf}.

\bibitem[Zhu et~al.(2024)Zhu, Yang, Wang, Song, Wu, Wei, and Li]{zhu2024pose}
Dawei Zhu, Nan Yang, Liang Wang, Yifan Song, Wenhao Wu, Furu Wei, and Sujian Li.
\newblock Po{SE}: Efficient context window extension of {LLM}s via positional skip-wise training.
\newblock In \emph{The Twelfth International Conference on Learning Representations}, 2024.
\newblock URL \url{https://openreview.net/forum?id=3Z1gxuAQrA}.

\end{thebibliography}
\bibliographystyle{neurips_data_2024}

\appendix

\section{Appendix}
\subsection{Additional Datasets}
\label{sec:add_data}
We list a few additional datasets as follows:

\textbf{GoEmotions}~\citep{demszky-etal-2020-goemotions} is the largest manually annotated dataset of 58k English comments from Reddit, which is labeled into 27 emotion categories or Neutral. There are 27 types of emotion types and drop the rare ones with few examples. Each selected example contains 28 tokens on average.

\textbf{Few-NERD}~\citep{ding-etal-2021-nerd} is a large-scale human-annotated name entity recognition dataset with a hierarchy of 8 coarse-grained and 66 fine-grained entity types. Each of the instances is a paragraph with approximately 61 tokens on average and contains one or multiple entity names as the ground truth answer. There are 66 types of entities in the collection.

The performance for the two tasks is demonstrated in Table~\ref{table:goemo_result_new} and Table~\ref{table:fewnerd_result}.

\begin{table}[tbh]
\small
\centering
\begin{tabular}{lccccccc}

\toprule
\multicolumn{1}{l}{\textbf{Model}} &\multicolumn{1}{c}{\textbf{Param}}
&\multicolumn{1}{c}{\textbf{Support}}  &\multicolumn{1}{c}{\textbf{1R}} &\multicolumn{1}{c}{\textbf{2R}}  &\multicolumn{1}{c}{\textbf{3R}}    &\multicolumn{1}{c}{\textbf{4R}}  &\multicolumn{1}{c}{\textbf{5R}}               \\
\midrule
\multicolumn{1}{l}{\textbf{Context Tokens}} &\multicolumn{1}{c}{\textbf{}}
&\multicolumn{1}{c}{\textbf{}}  &\multicolumn{1}{c}{\textbf{0.8K}} &\multicolumn{1}{c}{\textbf{1.6K}}  &\multicolumn{1}{c}{\textbf{2.4K}}    &\multicolumn{1}{c}{\textbf{3.2K}}  &\multicolumn{1}{c}{\textbf{4K}}  \\
\midrule

\multicolumn{1}{l}{Gemma-7B-base} & 7B & 8K & 0  & 0	& 0	& 0	& 0 \\

\multicolumn{1}{l}{LLaMA-2-7B-32K} & 7B & 32K & 0  & 0	& 0	& 0.2	& 0.2	\\

\multicolumn{1}{l}{ChatGLM3-6B-32K} & 6B & 32K & \textbf{22.0}  & 17.0	& 15.0	& 12.6	& 10.6  \\

\multicolumn{1}{l}{Qwen-1.5-7B-base} & 7B & 32K & 14.8  & \textbf{18.2}	& \textbf{18.6}	& \textbf{19.0}	& \textbf{14.2} \\

\multicolumn{1}{l}{Mistral-7B-v0.2-base} & 7B & 32K & 2.6  & 11.4	& 7.4	& 11.6	& 12.4 \\

\multicolumn{1}{l}{LLaMA-2-7B-LongLora} & 7B & 100K & 0  & 0	& 0	& 0	& 0 \\

\multicolumn{1}{l}{Yi-6B-200K} & 6B & 200K & 0  & 0	& 0.8	& 4.0	& 4.0 \\


\multicolumn{1}{l}{InternLM2-7B-base} & 7B & 200K & 0  & 0	& 0	& 0	& 0 \\

\multicolumn{1}{l}{Long-LLaMA-code-7B} & 7B & 256K & 0  & 0	& 0	& 0.2	& 0.4  \\

\midrule

\multicolumn{1}{l}{RWKV-5-World} & 7B & 4K & 8.8  & 7.4	& 4.6	& 5.2	& 4.0 \\

\multicolumn{1}{l}{Mamba-2.8B} & 2.8B & 2K & 0  & 0	& 0	& 0	& 0 \\
\midrule


\multicolumn{1}{l}{GPT4-turbo} & N/A & 128K & \textbf{36.5}  & \textbf{34.4}	&  \textbf{35.0}	&  \textbf{33.3}	& \textbf{32.0} \\

\multicolumn{1}{l}{GPT4o} & N/A & 128K & 23.0 & 23.8	&  21.2	&  21.2	& 22.2 \\

\multicolumn{1}{l}{Claude3-Opus} & N/A & 200K & 25.8  & 7.4	& 17.0	& 12.6	& 19.6 \\

\multicolumn{1}{l}{Gemini-1.5-Pro} & N/A & 10M & 19.0  & 10.4	& 9.2	&  10.6	&  9.4 \\

\midrule
\multicolumn{1}{l}{SoTA (BERT)} & N/A & - & \multicolumn{5}{c}{\textbf{58.9}} \\

\bottomrule
\end{tabular}
\caption{GoEmotion Result.}
\label{table:goemo_result_new}
\end{table}

\begin{table}[tbh]
\small
\centering
\begin{tabular}{lccccccc}

\toprule
\multicolumn{1}{l}{\textbf{Model}} &\multicolumn{1}{c}{\textbf{Param}}
&\multicolumn{1}{c}{\textbf{Support}}  &\multicolumn{1}{c}{\textbf{1R}} &\multicolumn{1}{c}{\textbf{2R}}  &\multicolumn{1}{c}{\textbf{3R}}    &\multicolumn{1}{c}{\textbf{4R}}  &\multicolumn{1}{c}{\textbf{5R}}               \\
\midrule
\multicolumn{1}{l}{\textbf{Context Tokens}} &\multicolumn{1}{c}{\textbf{}}
&\multicolumn{1}{c}{\textbf{}}  &\multicolumn{1}{c}{\textbf{5K}} &\multicolumn{1}{c}{\textbf{9K}}  &\multicolumn{1}{c}{\textbf{14K}}    &\multicolumn{1}{c}{\textbf{19K}}  &\multicolumn{1}{c}{\textbf{24K}} \\
\midrule

\multicolumn{1}{l}{Gemma-7B-base} & 7B & 8k & \textbf{44.0}  & 44.2	& 0	& 0	& 0 \\

\multicolumn{1}{l}{LLaMA-2-7B-32K} & 7B & 32k & 36.9  & 40.8	& 41.1	& 41.6	& 41.3	\\

\multicolumn{1}{l}{ChatGLM3-6B-32K} & 6B & 32k & 24.1  & 9.3	& 23.6	& 10.4	& 1.1  \\

\multicolumn{1}{l}{Qwen-1.5-7B-base} & 7B & 32k & 40.0  & 46.4	& 47.6	& 47.3	& 47.8 \\

\multicolumn{1}{l}{Mistral-7B-v0.2-base} & 7B & 32K & 42.2 & \textbf{47.4}  & \textbf{48.9}	& \textbf{50.0}	& \textbf{50.0} \\

\multicolumn{1}{l}{LLaMA-2-7B-LongLora} & 7B & 100K & 0  & 0	& 0	& 0	& 0 \\

\multicolumn{1}{l}{Yi-6B-200K} & 6B & 200k & 34.3  & 40.2	& 44.8	& 42.3	& 43.2 \\


\multicolumn{1}{l}{InternLM2-7B-base} & 7B & 200k & 43.6  & 46.2	& 46.5	& 47.8	& 48.3 \\

\multicolumn{1}{l}{Long-LLaMA-code-7B} & 7B & 256K & 22.3  & 25.5	& 26.5	& 29.4	& 27.0  \\

\midrule

\multicolumn{1}{l}{RWKV-5-World} & 7B & 1k & 13.9  & 0	& 0	& 0.7	& 9.9 \\

\multicolumn{1}{l}{Mamba-2.8B} & 2.8B & 2k & 0  & 0	& 0	& 0	& 0 \\
\midrule


\multicolumn{1}{l}{GPT4-turbo} & N/A & 128k & 53.4  & \textbf{55.3}	&  \textbf{56.2}	&  \textbf{55.6}	& \textbf{56.8} \\

\multicolumn{1}{l}{GPT4o} & N/A & 128k & 46.7  & 41.4	&  42.8 & 39.0 & 44.4 \\

\multicolumn{1}{l}{Claude3-Opus} & N/A & 200k & 53.5  & 51.3	&  51.2	&  52.4	&  52.5 \\

\multicolumn{1}{l}{Gemini-1.5-Pro} & N/A & 10M & \textbf{55.4}  & 47.8	& 49.5	&  41.4	&  42.4 \\

\midrule
\multicolumn{1}{l}{SoTA (PL-Marker)} & N/A & - & \multicolumn{5}{c}{\textbf{70.9}} \\

\bottomrule
\end{tabular}
\caption{Few-NERD Result.}
\label{table:fewnerd_result}
\end{table}

\subsection{Prompting Template}
\label{sec:prompt}
\begin{table}[tbh]
\small
\centering
\begin{tabular}{l p{12cm}}

\toprule
\multicolumn{1}{l}{\textbf{Dataset}} &\multicolumn{1}{c}{\textbf{Prompt}}  \\
\midrule

\multicolumn{1}{l}{GoEmotion} & Given a comment, please predict the emotion category of this comment. The prediction answer must come from the demonstration examples with the exact format. The examples are as follows: \newline \{comment: "...comment..."\newline emotion category: "...emotion..."\newline\} \textit{$\times$ repeat n times}\\

\midrule

\multicolumn{1}{l}{BANKING77} & Given a customer service query, please predict the intent of the query. The predicted answer must come from the demonstration examples with the exact format. The examples are as follows: \newline \{service query: "...service..."\newline intent category: "...intent..."\newline\} \textit{$\times$ repeat n times} \\

\midrule

\multicolumn{1}{l}{TacRED} & Given a sentence and a pair of subject and object entities within the sentence, please predict the relation between the given entities. The examples are as follows: \newline \{sentence: "...sentence... \newline the subject is "...subject..." \newline the object is "...object..." \newline the relation between the two entities is: "...relation..." \newline \} \textit{$\times$ repeat n times}\\

\midrule

\multicolumn{1}{l}{Few-NERD} & Given the sentence, please find the name entities in the sentence and their corresponding entity types in the strict format of the given examples as following (Entity: EntityType):\newline \{"...entity...": "...entity type..."\newline\} \textit{$\times$ repeat n times}\\

\midrule

\multicolumn{1}{l}{DialogRE} & Given the dialogue, please find the name pair entities in the dialogue and their corresponding relation types in the strict format of given examples as following (note that the number of entities has to strictly have the same value as the number of respective relation): \newline \{Dialogue: \newline "...dialogue..." \newline The list of entity pairs are "...(subject1, object1), (subject2, object2), etc... \newline The "...number of pairs..." respective relations between each entity pair are: "...relation, relation2, etc... \newline \} \textit{$\times$ repeat n times}\\

\midrule

\multicolumn{1}{l}{Discovery} & Given two sentence1 and sentence2, please predict the conjunction word between the two sentences. The predicted answer must come from the demonstration examples with the exact format. The examples are as follows: \newline \{"...sentence1..." ( ) "...sentence2..." \newline the conjunction word in ( ) is "...conjunction..." \newline \} \textit{$\times$ repeat n times}\\

\bottomrule
\end{tabular}
\caption{ The data prompt format of each dataset. Each dataset has a unique prompt format to effectively utilize the context and format of its respective data to get the best output response. }
\label{table:prompt}
\end{table}
The prompting template for each of the datasets is presented at Table~\ref{table:prompt}

\subsection{Additional Distribution Analysis}
\label{sec:group}

To facilitate a clear comparison between random and grouped distributions, we organize instances of the same class to be adjacent within the demonstration prompts. The impact of this reorganization on model performance, both pre and post-grouping, is presented in Table~\ref{table:label_instruction_result}.

\begin{table}[tbh]
\small
\centering
\begin{tabular}{lccccc}

\toprule
\multicolumn{1}{l}{\textbf{Model}} &\multicolumn{1}{c}{\textbf{Param}}
&\multicolumn{1}{c}{\textbf{Support}}  &\multicolumn{1}{c}{\textbf{Scatter}} &\multicolumn{1}{c}{\textbf{Grouped}}  &\multicolumn{1}{c}{\textbf{$\Delta$}}                \\
\midrule
\multicolumn{1}{l}{\textbf{Context Tokens}} &\multicolumn{1}{c}{\textbf{}}
&\multicolumn{1}{c}{\textbf{}}  &\multicolumn{3}{c}{\textbf{10K}}            \\
\midrule

\multicolumn{1}{l}{Gemma-7B-base} & 7B & 8K & 0  & 0 & 0	 \\

\multicolumn{1}{l}{LLaMA-2-7B-32K} & 7B & 32K & 0.4  & 3.0	& +2.6 \\

\multicolumn{1}{l}{ChatGLM3-6B-32K} & 6B & 32K & 38.9  & 35.6  & -3.3 \\

\multicolumn{1}{l}{Qwen-1.5-7B-base} & 7B & 32K & 45.2  & 33.0 & -12.2\\

\multicolumn{1}{l}{Mistral-7B-v0.2-base} & 7B & 32K & 51.6  & 5.1 & -46.5 \\

\multicolumn{1}{l}{LLaMA-2-7B-LongLora} & 7B & 100K & 0  & 0 & 0\\

\multicolumn{1}{l}{Yi-6B-200K} & 6B & 200K & 8.0  & 0 & -8 \\

\multicolumn{1}{l}{InternLM2-7B-base} & 7B & 200K & 15.5  & 4.8  & -9.7 \\

\multicolumn{1}{l}{Long-LLaMA-code-7B} & 7B & 256K & 4.1  & 0  & -4.1 \\

\midrule

\multicolumn{1}{l}{RWKV-5-World} & 7B & 4K & 1.0  & 3.6  & +2.6 \\

\multicolumn{1}{l}{Mamba-2.8B} & 2.8B & 2K & 0  & 0	& 0 \\

\midrule

\multicolumn{1}{l}{GPT4-turbo} & N/A & 128K   & 79.5  & 59.2	& -20.3 \\

\multicolumn{1}{l}{Gemini-1.5-Pro} & N/A & 10M & 79.6  & 57.3	& -22.3 \\

\bottomrule
\end{tabular}
\caption{Exploratory Result on TacRED 3 Round. \textbf{Grouped} means forcing the same-typed demonstration examples near by each other instead of randomly distributing in the prompt.}
\label{table:label_instruction_result}
\end{table}
The distribution plots for other models are shown in Figure~\ref{fig:group_ana_app1} and Figure~\ref{fig:group_ana_app2}.

\begin{figure}[th!]
\includegraphics[width=\textwidth]
{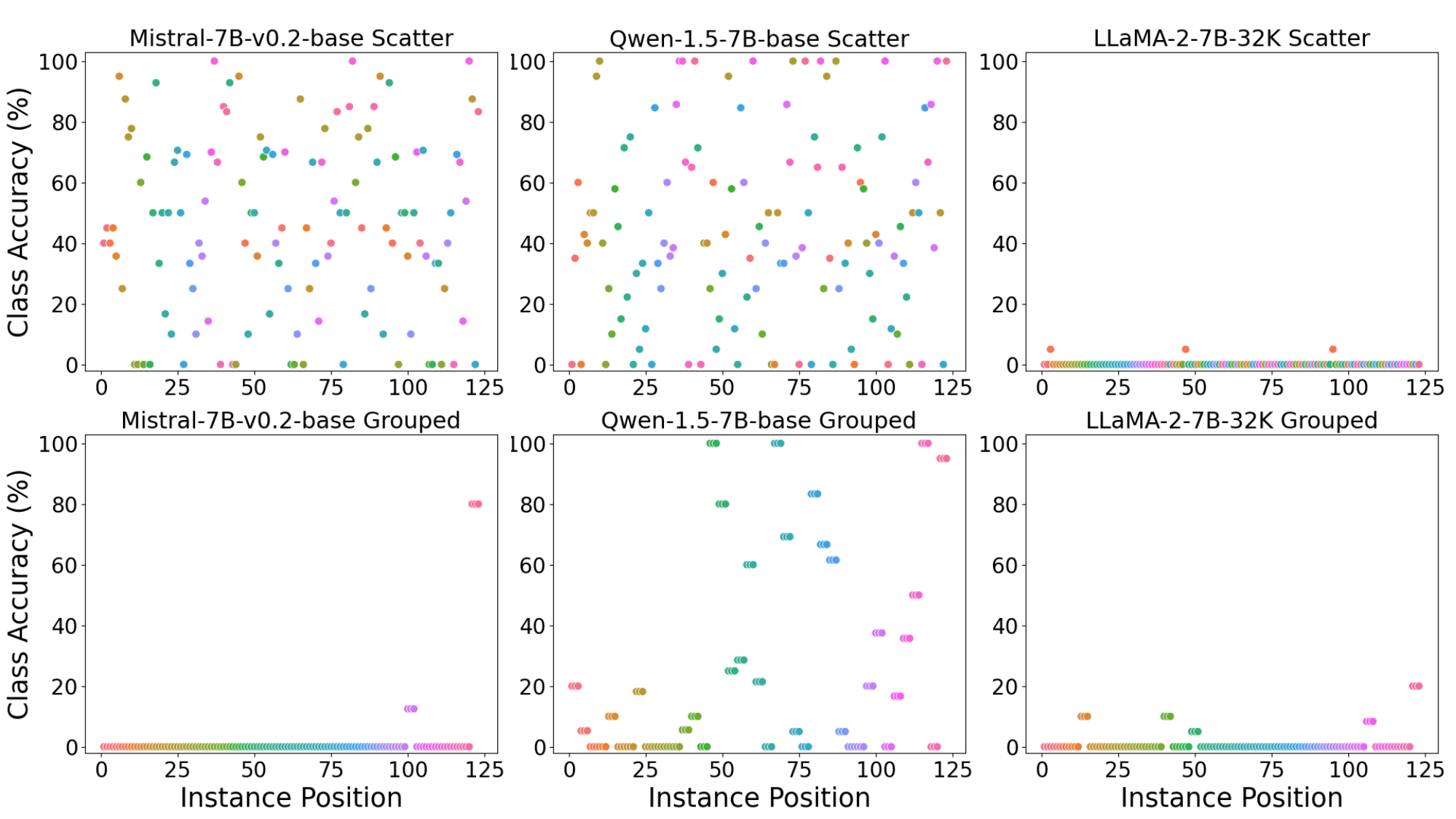}
  \caption{Visualization of accuracy for every class when instances from the same class are scattered V.S. grouped in the demonstration prompt.}
\label{fig:group_ana_app1}
\vspace{-3ex}
\end{figure}

\begin{figure}[th!]
\includegraphics[width=\textwidth]
{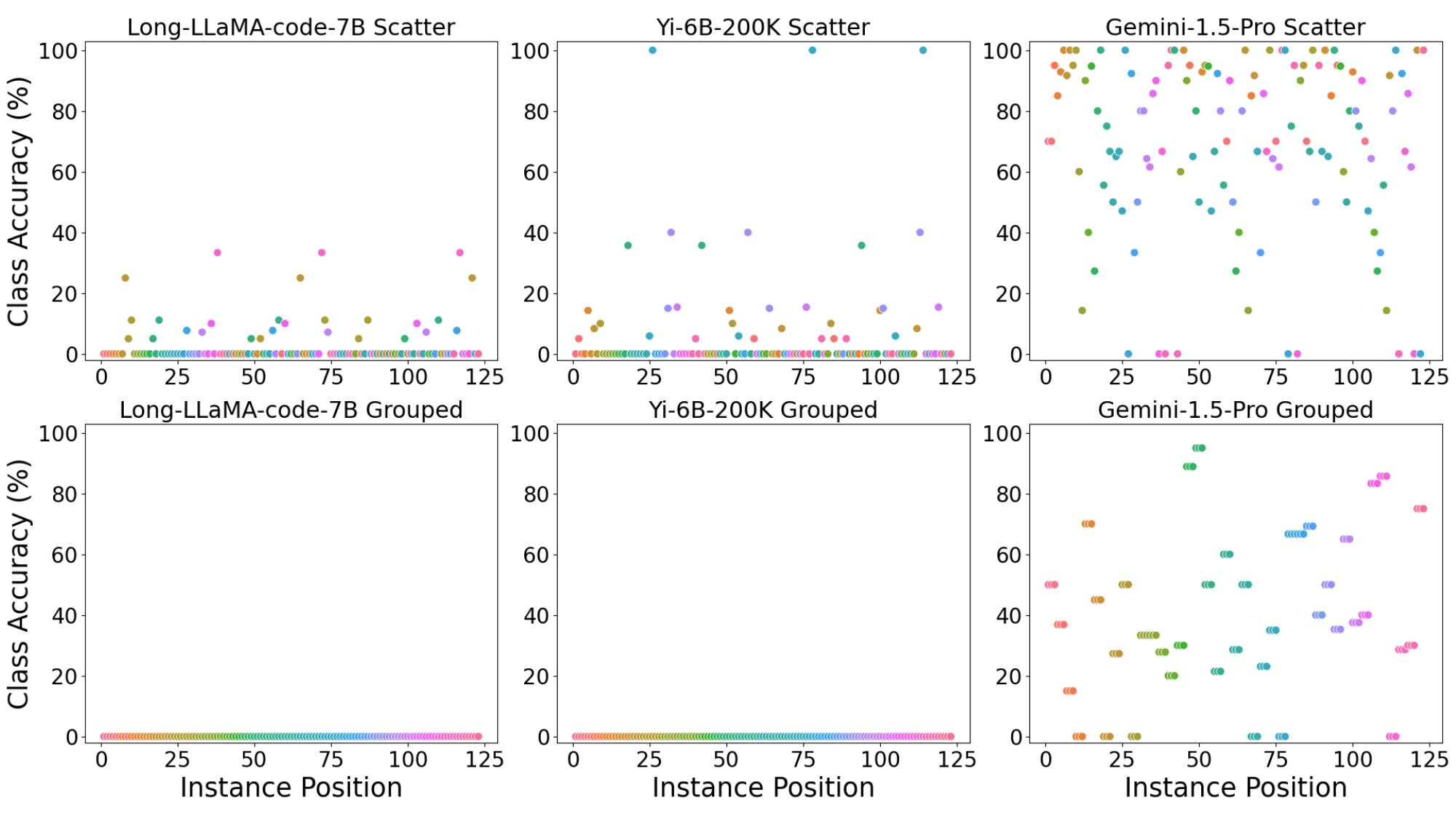}
  \caption{Visualization of accuracy for every class when instances from the same class are scattered V.S. grouped in the demonstration prompt.}
\label{fig:group_ana_app2}
\vspace{-3ex}
\end{figure}

\subsection{Data Accessibility}
\label{sec:acc_data}
Our LongICLBench is set under MIT license, thus permission is granted, free of charge, to any person obtaining a copy of this dataset and associated documentation files. The datasests are curated under the rules guaranteed by the original dataset. There is no personally identifiable or offensive content in the dataset.

\subsection{Broader Society Impacts}
\label{sec:impact}
The development of long LLMs evaluation benchmarks and the corresponding insights can boost the development of long-context techniques, which can revolutionize fields requiring deep contextual understanding, such as legal analysis, long-form journalistic content generation, and comprehensive academic summarization. However, there are potential risks associated with the deployment of such powerful models. Enhanced long-context capabilities could be misused for generating misinformation, especially in political or social contexts, where nuanced long-form content can have significant influence. There is also the risk of dependency on automated systems in critical decision-making processes, which could lead to over-reliance on technology at the expense of human judgment.

\subsection{Limitation}
\label{sec:limit}
While the proposed \eval can reveal the fact that current long LLMs still struggle with long-context understanding, it has limitations that \eval currently encompasses only one type of evaluation application, specifically on extreme-label classification with long in-context learning. The intricacies involved in other novel long-context tasks that require a full understanding of the whole long-context are waiting to be put forward.

\end{document}